%% file: main.tex
\documentclass{article} 
\usepackage{iclr2024_conference,times}

\input{math_commands.tex}

\usepackage{hyperref}
\usepackage{url}
\usepackage{adjustbox}
\usepackage{lipsum}
\usepackage{color}
\usepackage{wrapfig}
\usepackage{booktabs}
\usepackage{multirow,mathtools} 
\usepackage{algorithm,algpseudocode}
\usepackage{threeparttable}
\usepackage{amsthm}
\usepackage{amsmath}
\usepackage{amssymb}
\usepackage{booktabs}
\usepackage{amsfonts}
\usepackage{wrapfig}
\usepackage{times}
\usepackage{multirow}
\usepackage{url}
\usepackage{color}
\usepackage{xcolor}
\usepackage{tabularx}
\usepackage{caption}

\usepackage{enumitem}
\usepackage{subcaption}
\usepackage{natbib}
\usepackage{wrapfig}
\usepackage{resizegather}
\usepackage{algorithm}
\usepackage{algpseudocode}
\usepackage{bbm}
\usepackage{makecell}
\usepackage{wrapfig}
\usepackage{mdframed}
\usepackage[many]{tcolorbox}
\newtcolorbox{questionblock}{
    enhanced,
    boxrule=0pt,frame hidden,
    interior style={
        top color=lightgray,
        bottom color=lightgray
    },
    colback=lightgray,
    colframe=white,
    sharpish corners,
    borderline north={0pt,none,white},
    borderline south={0pt,none,white},
    borderline east={0pt,none,white},
    borderline west={0pt,none,white},
}

\usepackage{pifont}

\newcommand{\bdelta}{\boldsymbol\delta}
\newcommand{\bx}{\mathbf{x}}
\newcommand{\btheta}{\boldsymbol\theta}
\newcommand{\bmm}{\mathbf{m}}
\newcommand{\bc}{\mathbf{c}}

\title{Visual Prompting Upgrades Neural Network Sparsification: A Data-Model Perspective}

\author{Can Jin$^{1}$\thanks{Equal contribution.},\,
  Tianjin Huang$^{2,3}$\footnotemark[1],\,
  Yihua Zhang$^{4}$, 
  Mykola Pechenizkiy$^{2}$,
  Sijia Liu$^{4}$, \\
  \textbf{Shiwei Liu}$^{2,5}$, 
  \textbf{Tianlong Chen}$^{6,7,8}$ \\
  $^1$University of Science and Technology of China,
  $^2$Eindhoven University of Technology,\\
  $^3$University of Exeter,
  $^4$Michigan State University,
  $^5$University of Texas at Austin,\\
  $^6$MIT,
  $^7$University of North Carolina at Chapel Hill,
  $^8$Harvard University\\
  \texttt{jincan3@mail.ustc.edu.cn, t.huang2@exeter.ac.uk,} \\ \texttt{\{m.pechenizkiy,s.liu3\}@tue.nl, \{zhan1908,liusiji5\}@msu.edu}\\
  \texttt{tianlong@mit.edu}
}

\iclrfinalcopy 
\begin{document}

\maketitle

\begin{abstract}
\vspace{-2mm}
The rapid development of large-scale deep learning models questions the affordability of hardware platforms, which necessitates the pruning to reduce their computational and memory footprints. Sparse neural networks as the product, have demonstrated numerous favorable benefits like low complexity, undamaged generalization, \textit{etc}. Most of the prominent pruning strategies are invented from a \textit{model-centric} perspective, focusing on searching and preserving crucial weights by analyzing network topologies. However, the role of data and its interplay with model-centric pruning has remained relatively unexplored. In this research, we introduce a novel \textit{data-model co-design} perspective: to promote superior weight sparsity by learning important model topology and adequate input data in a synergetic manner. Specifically, customized \textbf{V}isual \textbf{P}rompts are mounted to upgrade neural \textbf{N}etwork \textbf{s}parsification in our proposed \textbf{\texttt{VPNs}} framework. As a pioneering effort, this paper conducts systematic investigations about the impact of different visual prompts on model pruning and suggests an effective joint optimization approach. Extensive experiments with $3$ network architectures and $8$ datasets evidence the substantial performance improvements from \textbf{\texttt{VPNs}} over existing start-of-the-art pruning algorithms. Furthermore, we find that subnetworks discovered by \textbf{\texttt{VPNs}} from pre-trained models enjoy better transferability across diverse downstream scenarios. These insights shed light on new promising possibilities of data-model co-designs for vision model sparsification. Code is available at \url{https://github.com/UNITES-Lab/VPNs}.

\end{abstract}

\vspace{-1mm}
\section{Introduction}
\vspace{-1mm}

Large-scale neural networks like vision and language models~\citep{brown2020language,radford2019language,touvron2023llama,chiang2023vicuna,li2022blip,bai2023qwen} have attracted stupendous attention in nowadays deep learning community, which pose significantly increased demands to computing resources. While remarkable performance has been offered, they suffer from prohibitively high training and inference costs, and the deployment of these gigantic models entails substantial memory and computational overhead. For instance, inferring the GPT-$3$ with $175$B parameters requires at least five $80$GB A$100$ GPUs~\citep{frantar2023massive}. 

To establish economic and lightweight network alternatives, model compression serves as an effective tool, gaining great popularity~\citep{dettmers2022llm,frantar2023massive,yao2022zeroquant,sun2023simple}. Among plenty of efforts for compression, model pruning~\citep{lecun1989optimal,gale2019state,frankle2018lottery,chen2020lottery} is one of the dominant approaches, aiming to trim down the least significant weights without hurting model performance. It is usually applied subsequent to the convergence of training~\citep{frankle2018lottery,chen2020lottery,molchanov2016pruning},  during the training process~\citep{zhu2017prune,gale2019state,chen2021chasing}, and even prior to the initiation of training~\citep{mocanu2018scalable,lee2018snip,evci2020rigging}. The resulting sparsity
ranges from fine-grained elements like individual weights~\citep{zhu2017prune} to coarse-grained structures such as neurons~\citep{hu2016network}, blocks~\citep{lagunas2021block}, filters~\citep{yin2023dynamic}, and attention heads~\citep{shim2021layer}. It is worth mentioning that the majority, if not all, of the conventional pruning algorithms, produce sparse neural networks in a \textit{model-centric} fashion -- analyzing architecture topologies and capturing their key subset by learning parameterized weight masks~\citep{sehwag2020hydra} or calculating proxy heuristics based on training dynamics~\citep{han2015deep}, architecture properties~\citep{hoang2023revisiting}, \textit{etc}.

Thanks to the recent advances in large language models (LLMs), the \textit{data-centric} AI regains a spotlight. Techniques like in-context learning~\citep{brown2020language,shin2020autoprompt,liu2022few} and prompting~\citep{liu2023pre,li2021prefix} construct well-designed prompts or input templates to empower LLMs and reach impressive performances on a variety of tasks. It evidences that such data-centric designs effectively extract and compose knowledge in learned models~\citep{wei2022emergent}, which might be a great assistant to locating critical model topologies. Nevertheless, the influence of data-centric methods on network sparsification has been less studied. To our best knowledge, only one concurrent work~\citep{xu2023compress} has explored the possibility of learning \textit{post-pruning prompts} to recover compressed LLMs. Thus, We focus on a different aim:
\vspace{-1mm}
\begin{center}
\textit{How to leverage prompts to upgrade vision model sparsification, from a data-model perspective?}
\end{center}
\vspace{-1mm}
Note that the effect of visual prompts on sparse vision models remains mysterious. Also, visual prompts are inherently more complex to comprehend and typically pose greater challenges in terms of both design and learning, in comparison to their linguistic counterparts.

To answer the above research questions, we start with a systematical pilot investigation of existing post-pruning prompts~\citep{xu2023compress} on sparse vision models. As presented in Section~\ref{sec: Pilot Study},~\textbf{directly inserting \textit{post-pruning} visual prompts into sparse vision models does not necessarily bring performance gains.} 
To unlock the capacity of visual prompts in sparse vision models, we propose a \textit{data-model co-design} paradigm. Specifically, we propose \textbf{\texttt{VPNs}} (\textbf{V}isual \textbf{P}rompting Upgrades \textbf{N}etworks \textbf{S}parsification) that co-trains the visual prompts with parameterized weight masks, exploring superior subnetworks. Our efforts are unfolded with the following five thrusts:
\vspace{-1mm}
\begin{itemize}[leftmargin=*]
\item [$\star$] (A Pilot Study) We conduct a pilot study of post-pruning prompts in sparse vision models and surprisingly find its inefficacy in improving the performance of well fine-tuned sparse vision models. 


\item [$\star$] (Algorithm) To unlock the potentials of visual prompts in vision model sparsification, we propose a novel \textit{data-model co-design} sparsification paradigm, termed \textbf{\texttt{VPNs}}, which simultaneously optimizes weight masks and tailored visual prompts.

\item [$\star$] (Experiments) We conduct extensive experiments across diverse datasets, architectures, and pruning regimes. Empirical results consistently highlight the impressive advancement of both performance and efficiency brought by \textbf{\texttt{VPNs}}. For example, \textbf{\texttt{VPNs}} outperforms the previous state-of-the-art (SoTA) methods \{HYDRA~\citep{sehwag2020hydra}, BiP~\citep{zhang2022advancing}, LTH~\citep{chen2021lottery}\} by \{$3.41\%$, $1.69\%$, $2.00\%$\} at $90\%$ sparsity with ResNet-$18$ on Tiny-ImageNet.

\item [$\star$] (Extra Findings) More interestingly, we demonstrate that the sparse masks from our \textbf{\texttt{VPNs}} enjoy enhanced transferability across multiple downstream tasks.
    
\item [$\star$] (Potential Practical Benefits) \textbf{\texttt{VPNs}} can be seamlessly integrated into structured pruning approaches, enabling more real-time speedups and memory reduction with competitive accuracies.
\end{itemize}

\vspace{-2mm}
\section{Related Works and A Pilot Study}
\vspace{-1mm}

\paragraph{Neural Network Pruning.} Pruning~\citep{mozer1989using, lecun1989optimal} aims at compressing networks by removing the least important parameters in order to benefit the model generalization~\citep{chen2022can}, robustness~\citep{sehwag2020hydra}, stability~\citep{hooker2020characterising}, transferability~\citep{chen2020lottery}, \textit{et al}. In the literature, an unpruned network is often termed the ``dense network", while its compressed counterpart is referred to as a ``subnetwork" of the dense network~\citep{chen2021lottery}. A commonly adopted compression strategy follows a three-phase pipeline: pre-training, pruning, and fine-tuning. Categorizing based on this pipeline, pruning algorithms can be segmented into post-training pruning, during-training pruning, and prior-training pruning. \textit{Post-training} pruning methods, applied after the dense network converges, are extensively explored. In general, these methods fall under three primary umbrellas: weight magnitude-based techniques~\citep{han2015deep}, gradient-centric methods~\citep{molchanov2016pruning, sanh2020movement, jiang2021towards}, and approaches leveraging Hessians~\citep{lecun1989optimal, hassibi1992second, dong2017learning}. Along with the rising of foundational models, more innovative post-training pruning methods have emerged to amplify their resource-efficiency~\citep{zafrir2021prune, peng2022length, lagunas2021block, frantar2021m, peng2021binary, peng2022towards, chen2022coarsening}. \textit{During-training} pruning, which is introduced by \citep{finnoff1993improving}, presents an effective variant for model sparsification. It begins by training a dense model and then iteratively trims it based on pre-defined criteria, until obtaining the desired sparse subnetwork. Significant contributions to this approach category are evident in works such as~\citep{gale2019state, chen2022sparsity, huang2022dynamic}. As a more intriguing yet challenging alternative, \textit{prior-training} pruning thrives~\citep{huang2023enhancing, jaiswal2023emergence}, which targets to identify the optimal subnetwork before fine-tuning the dense network. \citet{mocanu2018scalable, dettmers2019sparse, evci2020rigging} take a step further to advocate one particular group of sparse neural networks that are extracted at random initialized models, trained from the ground up, and able to reach commendable results.

\vspace{-2mm}
\paragraph{Prompting.} Traditionally, the quest for peak performance is centered on manipulating model weights. However, prompting heralds a pivotal shift towards \textit{data-centric} studies, illuminating the potential of innovative input design. The concept of prompting emerges in the domain of natural language processing (NLP) as a proficient approach for adapting pre-trained models to downstream tasks~\citep{liu2023pre}. Specifically, GPT-$3$ showcases its robustness and generalization to downstream transfer learning tasks when equipped with handpicked text prompts, especially in settings like few-shot or zero-shot learning~\citep{brown2020language}. There is a significant amount of work around refining text prompting, both in terms of crafting superior prompts~\citep{shin2020autoprompt, jiang2020can} and representing prompts as task-specific continuous vectors~\citep{liu2021p}. The latter involves optimizing these prompts using gradients during the fine-tuning phase, which is termed Prompt Tuning~\citep{li2021prefix, lester2021power, liu2021p}. Interestingly, this approach rivals the performance of full fine-tuning but enjoys the advantage of high parameter efficiency and low storage cost.
The design philosophy of prompt tuning is extended to the computer vision realm by~\citet{bahng2022exploring}, which incorporates prompt parameters directly into input images, thereby crafting a prompted image, termed Visual Prompt (VP). Building on this foundation, \citet{jia2022visual} proposes a visual-prompt tuning method that modifies pre-trained Vision Transformer models by integrating selected parameters into the Transformer's input space. \citet{chen2023understanding} reveals the importance of correct label mapping between the source and the target classes and introduces iterative label mapping to help boost the performance of VP. Further advancements are made by~\citet{liu2023explicit, zheng2022prompt, zhang2022neural}, which devise a prompt adapter towards enhancing or pinpointing an optimal prompt for a given domain. In a parallel approach, \citet{zang2022unified, zhou2022learning} and \citet{zhou2022conditional} introduce visual prompts in conjunction with text prompts to vision-language models, resulting in a noted improvement in downstream performance.

\vspace{-2mm}
\subsection{A Pilot Study}\label{sec: Pilot Study}
\vspace{-1mm}


\begin{wrapfigure}{r}{0.6\textwidth}
\vspace{-7mm}
\centering
  \begin{subfigure}{0.29\textwidth}
    \includegraphics[width=\linewidth]{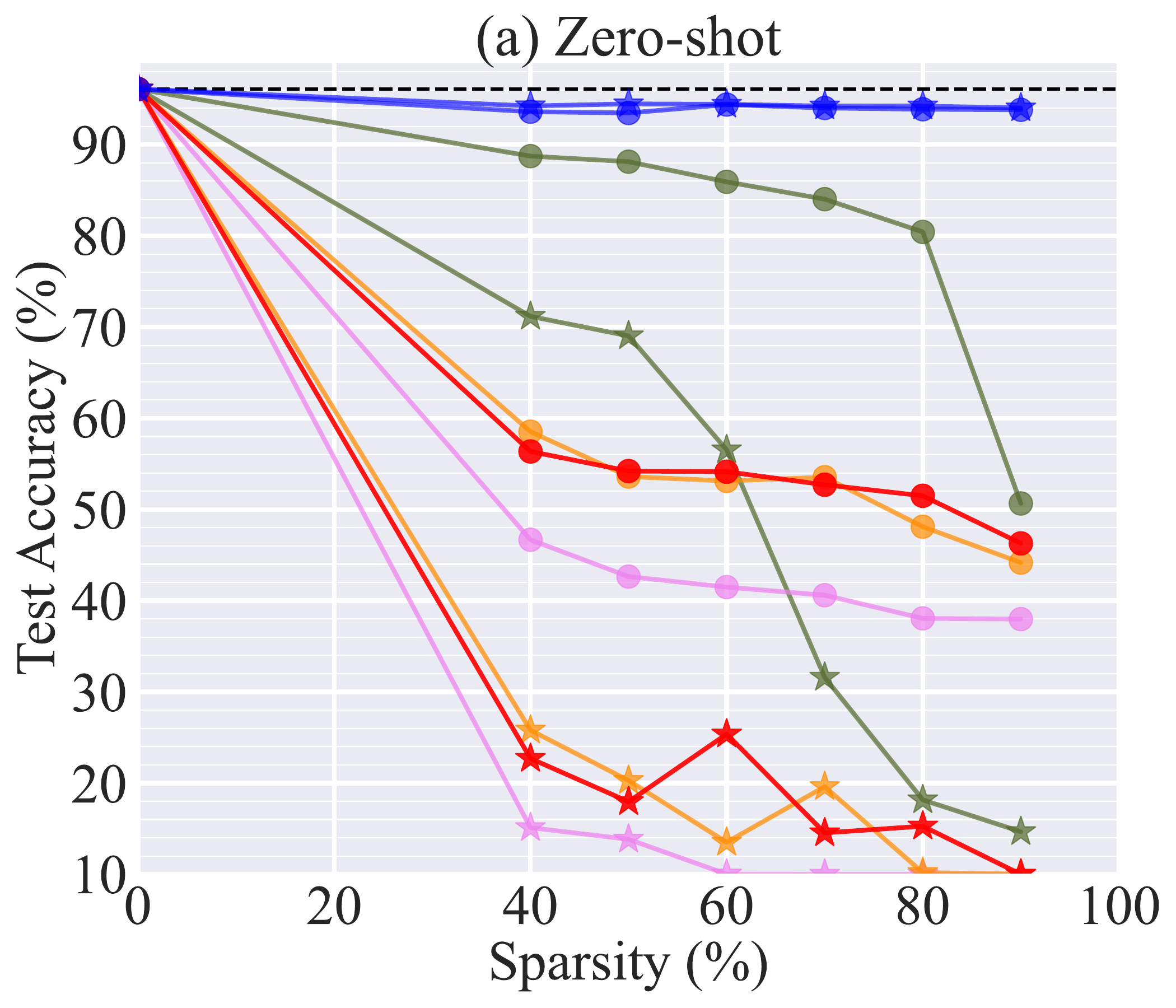}
  \end{subfigure}
  \hspace{0.1em}
  \begin{subfigure}{0.29\textwidth}
    \includegraphics[width=\linewidth]{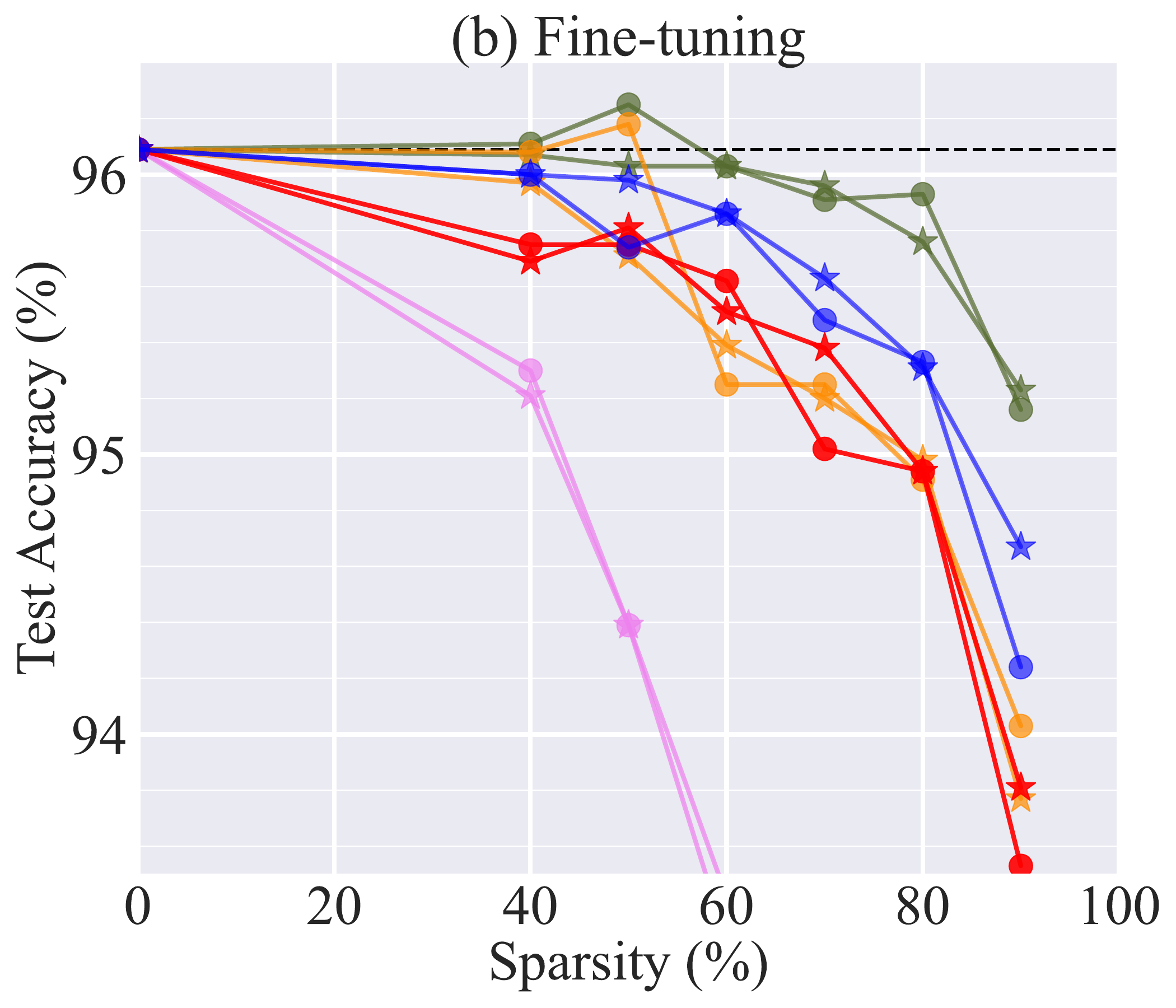}
  \end{subfigure}
  \begin{subfigure}{0.5\textwidth}
    \includegraphics[width=\linewidth]{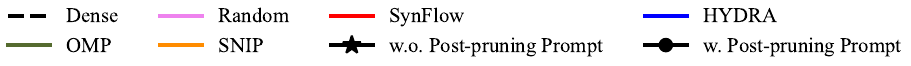}
  \end{subfigure}
\vspace{-3mm}
\caption{\small{\textbf{Post-pruning Prompt Results.} Performance of $5$ pruning methods and their post-pruning prompt counterparts on ResNet-$18$ and CIFAR$10$, which are marked as $\bullet$ and $\bigstar$, respectively. The dashed line indicates the dense network's performance. (a) Post-pruning with zero-shot. (b) Post-pruning with fine-tuning. \textit{Post-pruning prompt is only valid without fine-tuning.}}}
\label{Fig: Post-pruning Prompt}
\vspace{-5mm}
\end{wrapfigure}

\paragraph{Motivation.} The question of whether pruning should be either a more \textit{model-centric} or \textit{data-centric} process continues to be debated within the field. Certain proponents suggest pruning as \textit{model-centric}, with their assertions bolstered by the success of approaches like SynFlow \citep{tanaka2020pruning} which, despite not using any real data pass, deliver performances akin to dense networks. Yet, a considerable body of research contradicts this, emphasizing the superiority of post-training pruning techniques over prior-training ones, thereby articulating pruning's dependence on data~\citep{zhang2022advancing, liu2023sparsity}. To further complicate matters, the rise of LLMs has underscored the central role of data in shaping NLP's evolution. New strategies like in-context learning and prompting, designed to enhance LLMs' task-specific performance, have come to the fore. However, the precise role of \textit{data-centric} designs in sparsification remains an under-explored area, meriting further attention.

To the best of our knowledge, \citet{xu2023compress} is the sole concurrent study to delve into the potential of harnessing prompts to recover compressed LLMs. This research illuminates the efficacy of \textit{post-pruning} prompts, both manually crafted and learned ``soft" prompts, in enhancing the performance of compressed LLMs. However, the influence of VP on vision model sparsification presents an enigma, as VP is inherently more intricate and poses distinct challenges in designing and learning relative to their textual counterparts~\citep{bahng2022exploring}. To demystify it, we first investigate the post-pruning prompts on sparsified vision models. The experiments are conducted on ImageNet-$1$K pre-trained ResNet-$18$~\citep{he2016deep} and CIFAR$100$~\citep{krizhevsky2009learning}. We adopt $5$ pruning methods, \textit{i.e.}, Random~\citep{liu2022unreasonable}, OMP~\citep{han2015deep}, SNIP~\citep{lee2018snip}, SynFlow~\citep{tanaka2020pruning}, and HYDRA~\citep{sehwag2020hydra}, to analyze the performance of post-pruning prompts across various sparsity levels. To make a holistic study, we apply the post-pruning prompt to the sparse models with and without fine-tuning the subnetwork, referred to as \textbf{``Zero-shot''} and \textbf{``Fine-tuning''}, respectively. As shown in Figure~\ref{Fig: Post-pruning Prompt}, we find that: Post-pruning prompts only escalate the subnetworks before fine-tuning and bring marginal gains to the subnetwork with fine-tuning. The reason is likely that, after fine-tuning, the sparse model is sufficiently strong, leaving less room for prompts to enhance its performance. Neither of these settings consistently surpasses the standard no-prompting approach, which involves pruning and fine-tuning.


\vspace{-2mm}
\paragraph{Open Question.} As deliberated, the post-pruning prompting paradigm falls short in improving sparse vision models. This situation compels us to ask -- \textit{how to effectively utilize visual prompts to enhance the sparsification of vision models?} Our answer: a  \textbf{data-model co-design}  paradigm.

\vspace{-2mm}
\section{Methodology}
\vspace{-1mm}

\begin{figure}[t]
\vspace{-1mm}
\centering
\includegraphics[width=1\linewidth]{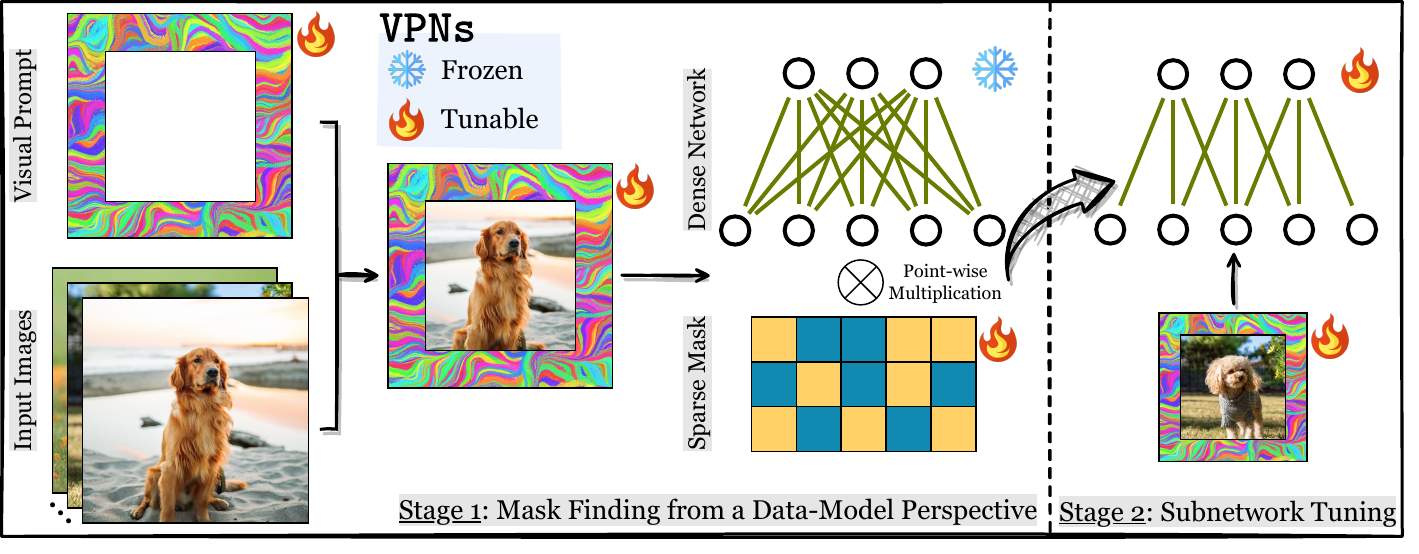}
\vspace{-7mm}
\captionof{figure}{\small{Overview of \textbf{\texttt{VPNs}}. In stage $1$, it locates superior sparse topologies from a data-model perspective. A tailored VP is added to input samples and weight masks are jointly optimized together with the VP. In stage $2$, the identified subnetwork is further fine-tuned with its VP.}}
\label{Fig: Our method}
\vspace{-3mm}
\end{figure}

In this section, we provide details about \textbf{\texttt{VPNs}}, which contains ($1$) designing appropriate visual prompts and ($2$) incorporating VPs to upgrade the sparse training of vision models in a \textbf{data-model} jointly optimization manner. An overview of our proposed \textbf{\texttt{VPNs}} is depicted in Figure~\ref{Fig: Our method}.

\vspace{-2mm}
\subsection{Designing Appropriate Visual Prompts}
\vspace{-1mm}
Visual prompts are proposed to address the problem of adapting a pre-trained source model to downstream tasks without any task-specific model modification, \textit{e.g.} fine-tuning network weights. To be specific, VP modifies the input image by injecting a small number of learnable parameters. Let $\mathcal{D} = \{(\bx_1, y_1),...,(\bx_n, y_n)\}$ denotes the vanilla downstream image dataset, $\bx$ is an original image in $\mathcal{D}$ with $y$ as its label, and $n$ represents the total number of images. The generic form of input prompting is then formulated as:
{\small\begin{equation}
    \bx'(\bdelta)=h(\bx, \bdelta), \bx \in \mathcal{D}= \{(\bx_1, y_1),...,(\bx_n, y_n)\},
    \label{Formula: Input Prompting}
\end{equation}}
\begin{wrapfigure}{r}{0.18\textwidth}
\centering
\includegraphics[width=\linewidth]{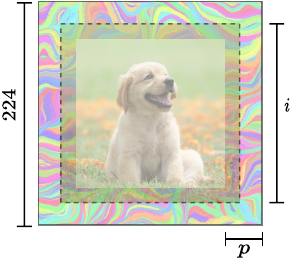}
\vspace{-5mm}
\caption{\small{Our VP}.}
\label{Fig: pad Prompt}
\vspace{-5mm}
\end{wrapfigure}
where $h(\cdot,\cdot)$ is an input transformation that integrates $\bx$ with the learnable input perturbation $\bdelta$ and $\bx'$ is the modified data after prompting. 

Our VP design \underline{first} resizes the original image $\bx$ to a specific \textbf{input size} $i\times i$ and pad it to $224\times224$ with $0$ values to get the resized image. We mark this process as $r^i(\bx)$, where $r(\cdot)$ refers to the resize and pad operation and $i$ indicates the target size, \textit{i.e.} input size. \underline{Subsequently}, we initiate the perturbation parameters of $\bdelta$ as a $224\times224$ matrix and mask a portion of them. Different visual prompts can be crafted by masking parameters in diverse shapes, locations, and sizes. In our case, the fixed mask is a central square matrix and the left four peripheral segments stay tunable. This kind of perturbation design is similar to \textit{pad prompt} in~\citet{bahng2022exploring} and the width of each peripheral side marked as $p$ is called \textbf{pad size}. More details about the prompt can be found in Appendix~\ref{Appendix: VP Design Details}. \underline{Finally}, the input prompting operation of \textbf{\texttt{VPNs}} is described as below (Figure~\ref{Fig: pad Prompt}):
{\small\begin{equation}
    \bx'(\bdelta) = h(\bx, \bdelta) = r^i(\bx) + \bdelta^p, \bx \in \mathcal{D},
    \label{Formula: Our Input Prompting}
\end{equation}}%
where $\bdelta^p$ is the pad prompt perturbation with a pad size of $p$. Note that, usually, $i+2p$ is larger than the input sample size like $224$ to sufficiently utilize all sample pixels.
\vspace{-2mm}
\subsection{Upgrading Network Sparsification with Visual Prompts}
\vspace{-1mm}
Given the input prompt formulation (Equation~\ref{Formula: Our Input Prompting}), VP seeks to advance downstream task performance of a pre-trained source model $f_{\btheta_{\text{pre}}}$ by optimizing the tunable part in $\bdelta$. Here $\btheta_{\text{pre}}$ refers to the pre-trained weights that are fixed in this stage. It raises a \textit{prompt optimization problem} as follows:
{\small\begin{equation}
    \underset{\bdelta}{\text{minimize}}\quad \mathbb{E}_{(\bx, y)\in \mathcal{D}}\mathcal{L}(f_{\btheta_{\text{pre}}}(\bx'(\bdelta)), y),
    \label{Formula: Prompt Optimization Problem}
\end{equation}}
where $\mathcal{L}$ is the objective function such as a cross-entropy loss for image recognition problems. As for the network sparsification, we recast it as an empirical risk minimization with respect to a learnable parameterized mask and the corresponding model weights can be frozen. Then a \textit{mask finding problem} is depicted as below:
{\small\begin{equation}
    \underset{\bmm}{\text{minimize}}
    \quad \mathbb{E}_{(\bx, y)\in \mathcal{D}}\mathcal{L}(f_{\btheta_{\text{pre}} \odot \bmm}(\bx), y),\quad s.t.\  ||\bmm||_{0}\leq (1-s)|\btheta_{\text{pre}}|,
    \label{Formula: Mask Finding Problem}
\end{equation}}%
where $\bmm$ is the mask variable, $\btheta_{\text{pre}} \odot \bmm$ is a point-wise multiplication between the mask and model weights, $s$ denotes the desired sparsity level, and $|\btheta_{\text{pre}}|$ refers to the number of parameters in $\btheta_{\text{pre}}$. 

Our proposed \textbf{\texttt{VPNs}} leverages visual prompts to upgrade the process of model sparsification by seamlessly integrating Equations~\ref{Formula: Prompt Optimization Problem} and~\ref{Formula: Mask Finding Problem}. To be specific, the joint optimization of prompt $\bdelta$ and $\bmm$ is described as follows:
{\small\begin{equation}
    \underset{\bmm,\bdelta}{\text{minimize}}\quad \mathbb{E}_{(\bx, y)\in \mathcal{D}}\mathcal{L}(f_{\btheta_{\text{pre}} \odot \bmm}(\bx'(\bdelta)), y)\quad s.t.\  ||\bmm||_{0}\leq (1-s)|\btheta_{\text{pre}}|,
    \label{Formula: VP Mask Finding Problem}
\end{equation}}%
where the learned mask $\bmm$ will be turned into a binary matrix in the end. The thresholding technique from~\citet{ramanujan2020s} is applied to map large and small scores to $1$ and $0$, respectively.

After obtaining sparse subnetworks from \textbf{\texttt{VPNs}}, a subsequent retraining phase is attached. It is another data-model co-optimization problem of VP and model weights as below: 
{\small\begin{equation}
    \underset{\bdelta,\btheta}{\text{minimize}}\quad \mathbb{E}_{(\bx, y)\in \mathcal{D}}\mathcal{L}(f_{\btheta \odot \bmm}(\bx'(\bdelta)), y)\quad s.t.\  \bmm=\bmm_{s},
    \label{Formula: VP Weight Tuning Problem}
\end{equation}}%
where $\btheta$ is the model parameters that are initialized as $\btheta_{\text{pre}}$. $\bmm_{s}$ represents the mask found by Equation~\ref{Formula: VP Mask Finding Problem}, and is fixed in this stage. 
\vspace{-2mm}
\subsection{Overall Procedure of \texttt{VPNs}}
\vspace{-1mm}
Our \textbf{\texttt{VPNs}} \underline{first} creates a VP following the Equation~\ref{Formula: Our Input Prompting}. \underline{Then}, to locate the \textbf{\texttt{VPNs}} sparse subnetwork, VP and the parameterized mask are jointly optimized based on Equation~\ref{Formula: VP Mask Finding Problem}. In this step, $\bmm$ is initialized with a scaled-initialization from~\citet{sehwag2020hydra}, $\bdelta$ adopts a $0$ initialization, and $\btheta$ is initialized with $\btheta_{\text{pre}}$ which stays frozen. \underline{Finally}, the weights of found sparse subnetwork are further fine-tuned together with VP, as indicated in Equation~\ref{Formula: VP Weight Tuning Problem}. During this step, $\btheta$ is initialized with $\btheta_{\text{pre}}$, visual prompt $\bdelta$ and mask $\bmm$ inherit the value of $\bdelta_s$ and $\bmm_{s}$ from the previous stage, respectively. Note that here $\bmm$ is kept frozen. The detailed procedure of \textbf{\texttt{VPNs}} is summarized in the Appendix~\ref{Appendix: VPNs Algorithm}. It is worth mentioning that such \textit{data-model} co-design, \textit{i.e.}, \textbf{\texttt{VPNs}}, presents a greatly improved efficiency in terms of searching desired high-quality subnetworks. For instance, compared to previous \textit{model-centric} approaches, \textbf{\texttt{VPNs}} only needs half the epochs of HYDRA~\citep{sehwag2020hydra} and OMP~\citep{han2015deep}, while achieving even better accuracy (see Table~\ref{Tab. Epochs and Steps}).
\vspace{-2mm}
\section{Experiments}
\vspace{-1mm}
To evaluate the effectiveness of our prompting-driven sparsification method, we follow the most common evaluation of visual prompting, i.e., evaluating sparse models pre-trained on a large dataset (ImageNet-1K) on various visual domains. Moreover, we conduct extensive empirical experiments including ($1$) Affirming the superior performance of \textbf{\texttt{VPNs}} over different datasets and architectures; ($2$) The transferability of \textbf{\texttt{VPNs}} across different datasets is investigated; ($3$) We further analyze the computational complexity of \textbf{\texttt{VPNs}} through the lens of time consumption, training epochs, and gradient calculating steps; ($4$) Our study also encompasses in-depth investigations into structured pruning algorithms; ($5$) Ablation studies are presented, which concentrate on the influence of different VP methods, pad sizes, and input sizes. 
\vspace{-2mm}
\subsection{Implementation Details}\label{sec. implementation details}
\vspace{-1mm}
\paragraph{Network and Datasets.}
We use three pre-trained network architectures for our experiments -- ResNet-$18$~\citep{he2016deep}, ResNet-$50$~\citep{he2016deep}, and VGG-$16$~\citep{simonyan2014very}, which can be downloaded from official Pytorch Model Zoo\footnote{\href{https://pytorch.org/vision/stable/models.html}{https://pytorch.org/vision/stable/models.html}}. These models are pre-trained on the ImageNet-$1$K dataset \citep{deng2009imagenet}. We then evaluate the effectiveness of VPNs over \textbf{eight} downstream datasets -- Tiny ImageNet~\citep{le2015tiny}, StanfordCars~\citep{krause20133d}, OxfordPets~\citep{parkhi2012cats}, Food$101$~\citep{bossard2014food}, DTD~\citep{cimpoi2014describing}, Flowers$102$~\citep{nilsback2008automated}, CIFAR$10$/$100$~\citep{krizhevsky2009learning}, respectively. Further details of the datasets can be found in Table~\ref{Tab: Appendix Dataset Attributes}. 

\vspace{-2mm}
\paragraph{Pruning Baselines.} 
We select \textbf{eight} representative state-of-the-art (SoTA) pruning algorithms as our baselines. ($1$) \textit{Random Pruning} (Random) \citep{liu2022unreasonable} is commonly used as a basic sanity check in pruning studies. ($2$) \textit{One-shot Magnitude Pruning} (OMP) removes weights with the globally smallest magnitudes \citep{han2015deep}. ($3$) \textit{The lottery ticket hypothesis} (LTH) iteratively prunes the $20\%$ of remaining weights with the globally least magnitudes and rewinds model weights to their initial state. In our experiments, weights are rewound to their ImageNet-$1$K pre-trained weights, following the default configurations in~\citet{chen2021lottery}.  ($4$) \textit{Pruning at initialization} (PaI) locates sparse subnetworks at the initialization phase by the defined salience metric. We opt for three widely-recognized methodologies: SNIP \citep{lee2018snip}, GraSP \citep{wang2020gradient}, and SynFlow \citep{tanaka2020pruning} ($5$) \textit{HYDRA} \citep{sehwag2020hydra} prunes weights based on the least importance scores, which is the most important baseline as it can be seen as our method without the visual prompt design. ($6$) \textit{BiP} \citep{zhang2022advancing}, characterized as a SoTA pruning algorithm, formalizes the pruning process within a bi-level optimization framework.

\vspace{-2mm}
\paragraph{Training and Evaluation.}
We follow the pruning baselines implementation in \citep{liu2022unreasonable}, selecting optimal hyper-parameters for various pruning algorithms by grid search. As for visual prompts, our default VP design in \textbf{\texttt{VPNs}} employs a pad prompt with an input size of $224$ and a pad size of $16$. We also use an input size of 224 for all the baselines to ensure a fair comparison. More implementation details are in Table~\ref{Tab: Appendix Hyperparameters and Configurations}.

\begin{figure}[!ht]
\vspace{-3mm}
  \centering
  \begin{subfigure}{0.24\textwidth}
    \includegraphics[width=\linewidth]{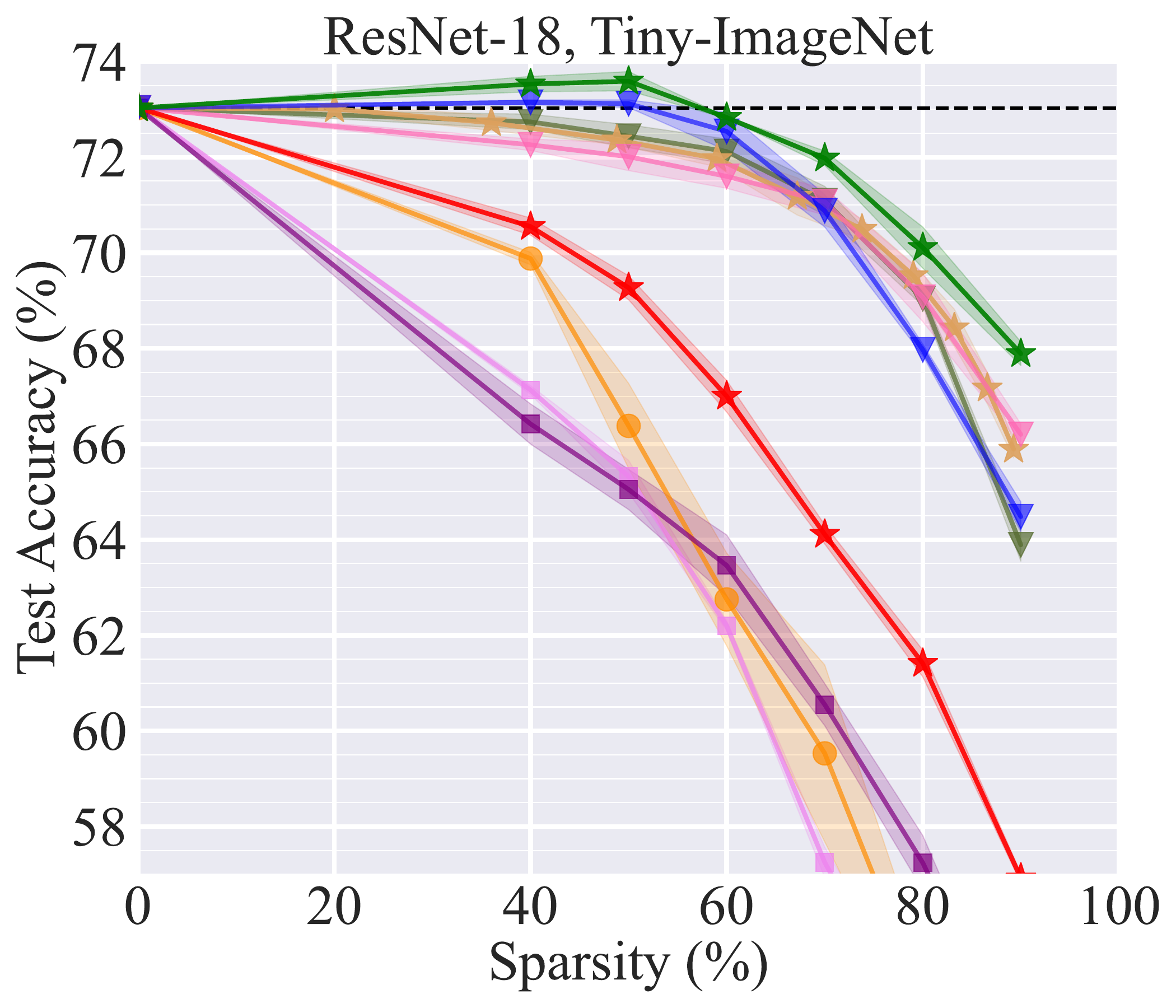}
  \end{subfigure}
  \begin{subfigure}{0.24\textwidth}
    \includegraphics[width=\linewidth]{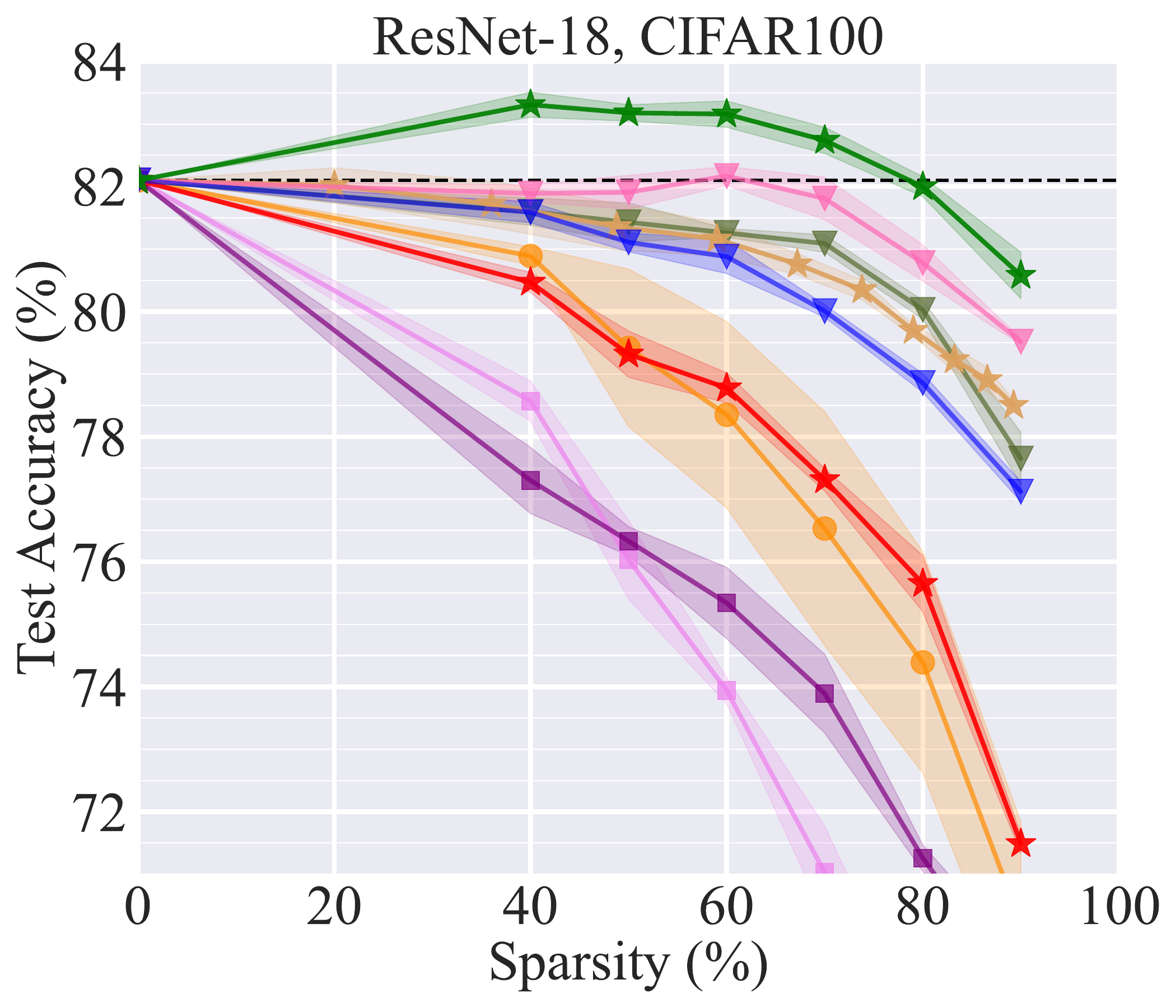}
  \end{subfigure}
  \begin{subfigure}{0.24\textwidth}
    \includegraphics[width=\linewidth]{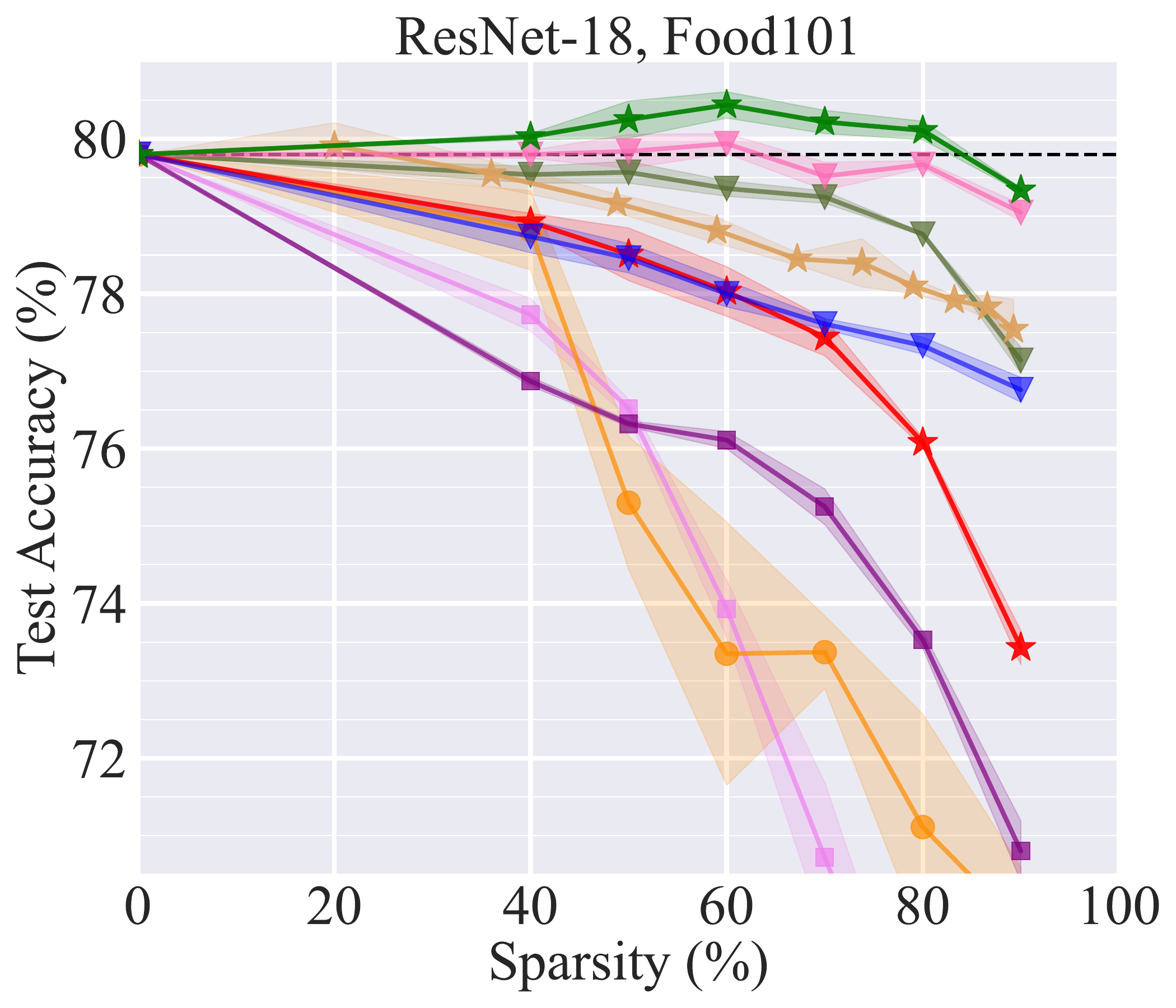}
  \end{subfigure}
  \begin{subfigure}{0.24\textwidth}
    \includegraphics[width=\linewidth]{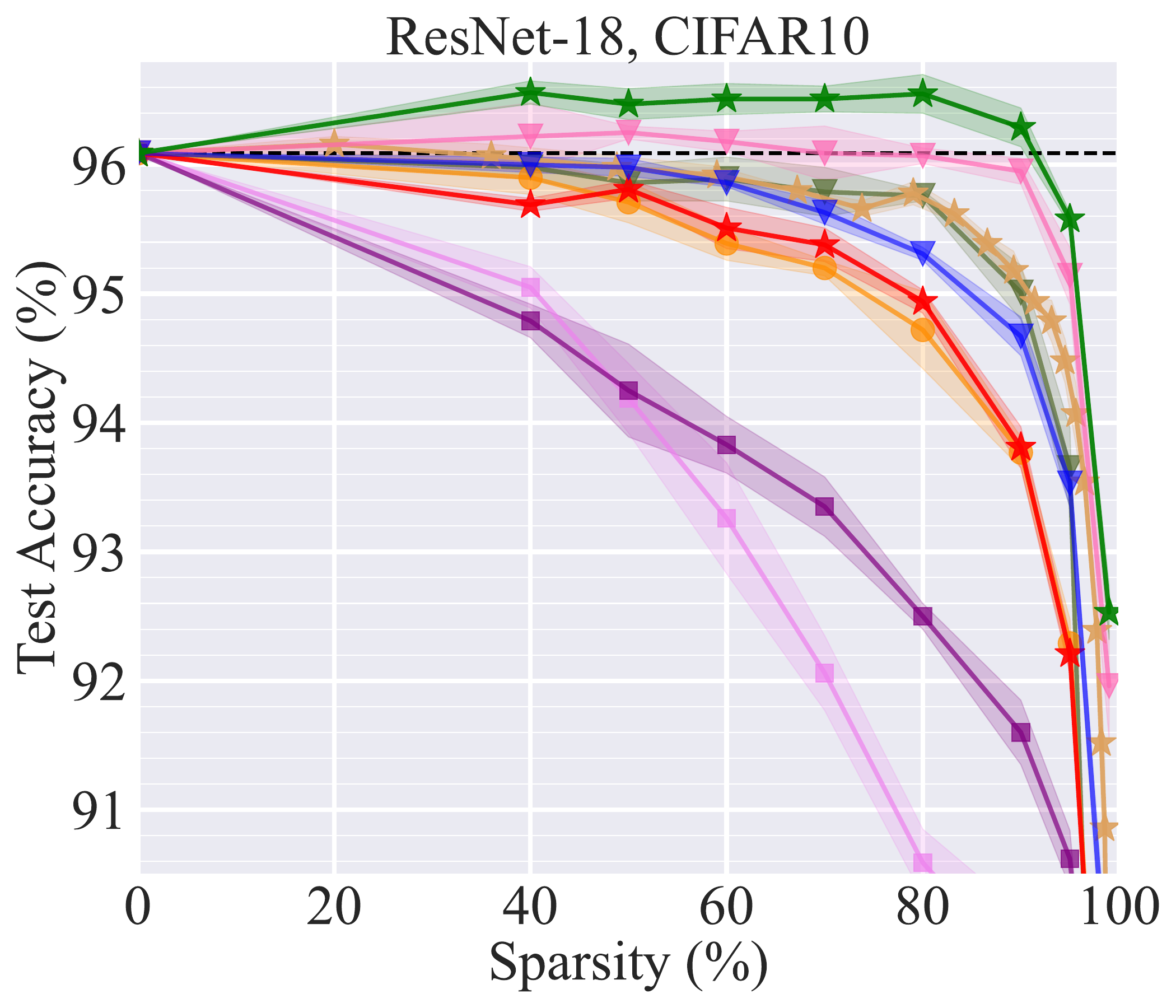}
  \end{subfigure}

  \begin{subfigure}{0.24\textwidth}
    \includegraphics[width=\linewidth]{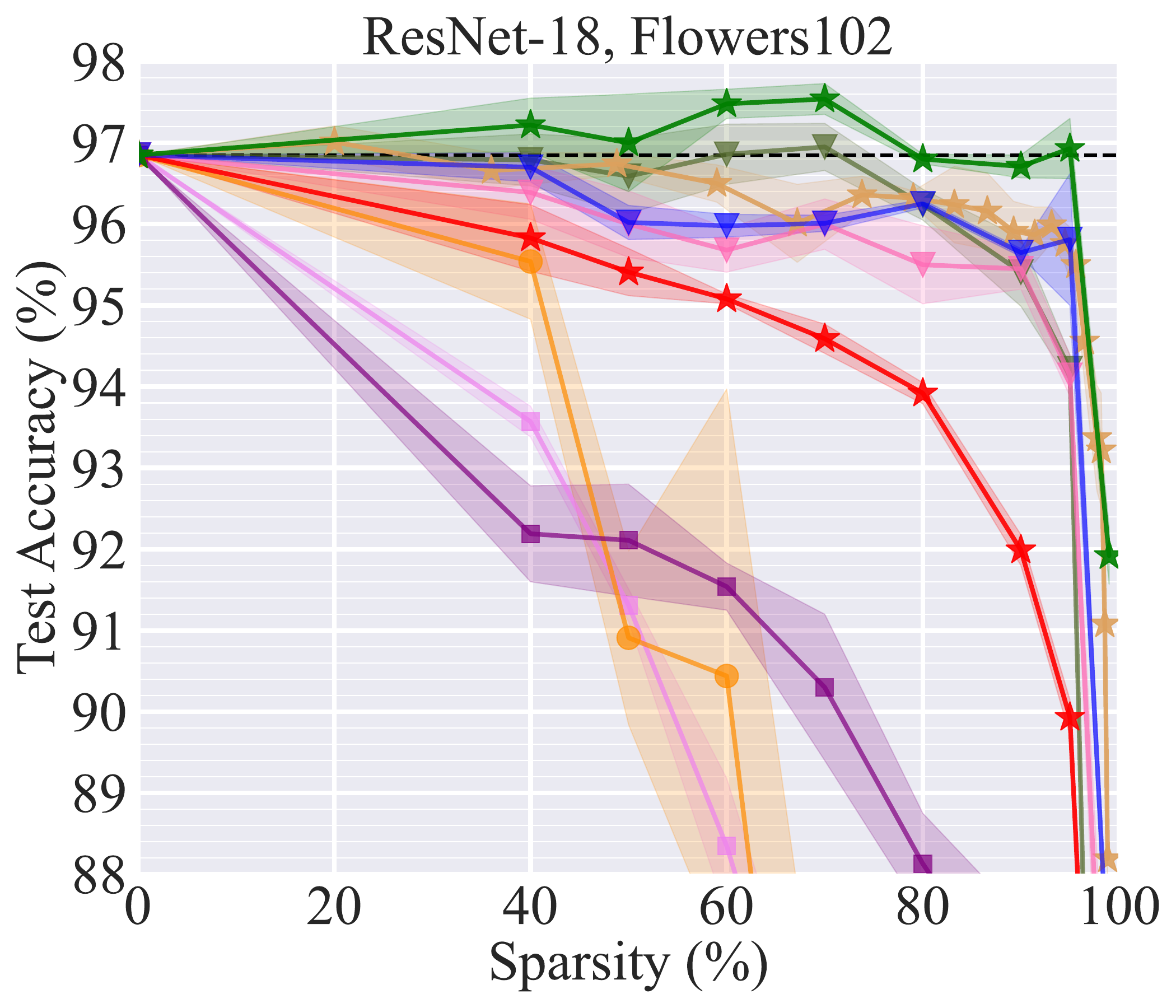}
  \end{subfigure}
  \begin{subfigure}{0.24\textwidth}
    \includegraphics[width=\linewidth]{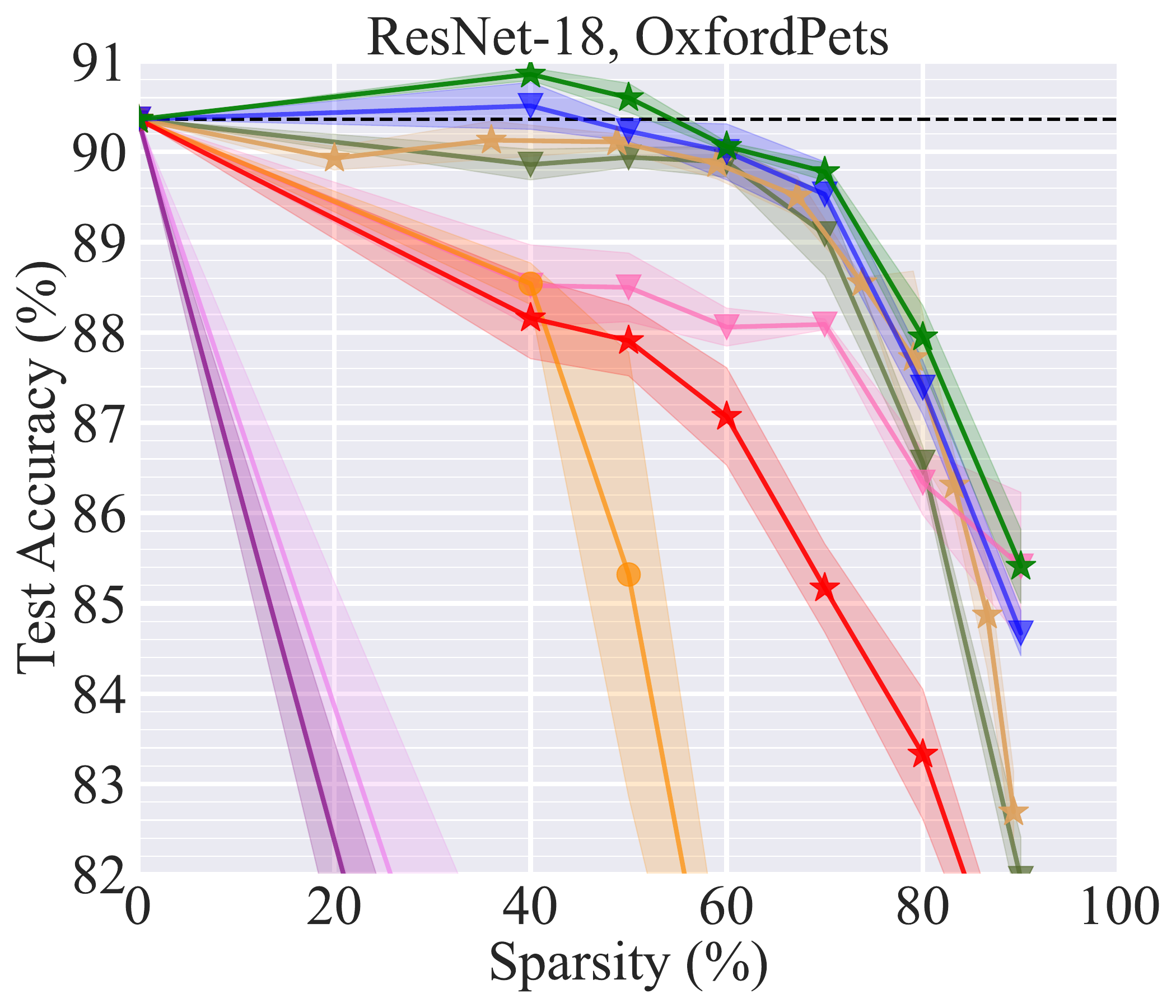}
  \end{subfigure}
  \begin{subfigure}{0.24\textwidth}
    \includegraphics[width=\linewidth]{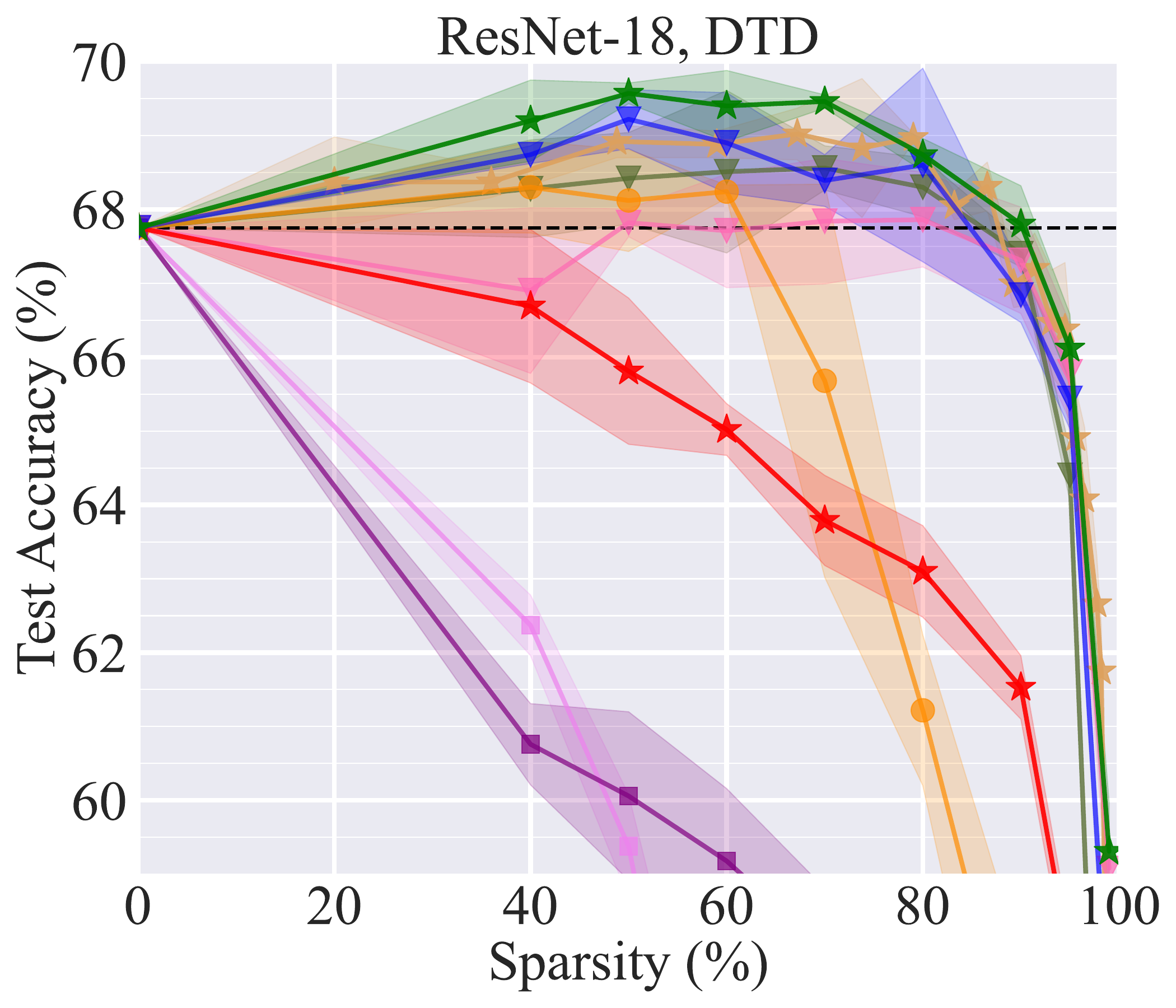}
  \end{subfigure}
  \begin{subfigure}{0.24\textwidth}
    \includegraphics[width=\linewidth]{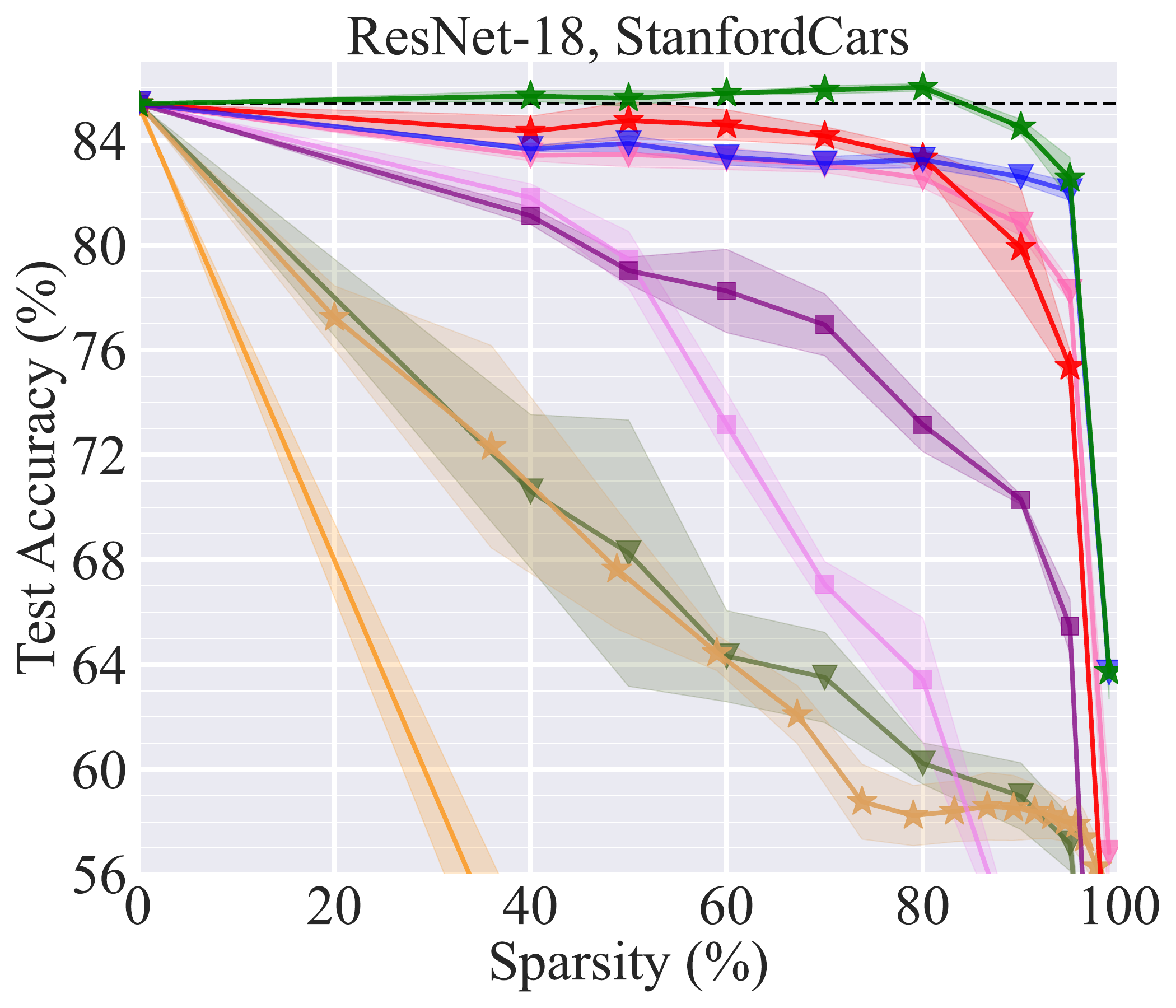}
  \end{subfigure}
  \begin{subfigure}{0.55\textwidth}
    \includegraphics[width=\linewidth]{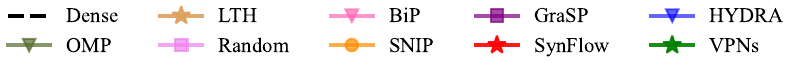}
  \end{subfigure}
  \vspace{-3mm}
  \caption{\small{\textbf{Downstream Fine-tuning Results.} The performance overview of $9$ unstructured pruning algorithms. All the models are pre-trained on ImageNet-1K; and then pruned and fine-tuned both on the specific downstream dataset. The performance of the dense model and \texttt{VPNs}' best are marked using dashed lines. All the results are averaged over 3 runs. \texttt{VPNs} consistently outperforms other baselines on all \textbf{eight} tasks.}}
  \label{Fig. Superior Performance}
  \vspace{-2mm}
\end{figure}

\vspace{-3mm}
\subsection{Main Results}
\vspace{-1mm}
\paragraph{Superior Performance of \texttt{VPNs}.}
Using a ResNet-$18$ pre-trained on ImageNet-$1$K, we evaluate the capability of \texttt{VPNs} in pruning models across multiple downstream datasets. As illustrated in Figure~\ref{Fig. Superior Performance}, several positive observations can be drawn: \ding{182} The dominance of \texttt{VPNs} is especially pronounced on larger datasets such as Tiny-ImageNet, CIFAR$100$, Food$101$, and CIFAR$10$. At $90\%$ sparsity level, \texttt{VPNs} outperforms \{HYDRA, BiP, LTH\} by \{$3.41\%$, $1.69\%$, $2.00\%$\} on Tiny-ImageNet and surpasses \{HYDRA, BiP, OMP\} by \{$3.46\%$, $2.06\%$, $2.93\%$\} on CIFAR$100$. \ding{183} \texttt{VPNs} still delivers top-tier results on smaller datasets like Flowers$102$, OxfordPets, DTD, and StanfordCars. For instance, the test accuracy of \texttt{VPNs} is \{$1.12\%$, $2.79\%$, $2.71\%$\} higher than \{HYDRA, BiP, OMP\} at $95\%$ sparsity on Flowers$102$. \ding{184} \texttt{VPNs} outperforms fully fine-tuned dense models at high sparsity levels on all eight downstream datasets. It finds subnetworks better than dense counterparts at \{$50\%$, $70\%$, $80\%$, $90\%$\} sparsity on \{Tiny-ImageNet, CIFAR$100$, Food$101$, CIFAR$10$\} and \{$70\%$, $50\%$, $90\%$, $90\%$\} sparsity on \{Flowers$102$, OxfordPets, DTD, StanfordCars\}. 

\begin{figure}[t]
\vspace{-2mm}
  \centering
  \begin{subfigure}{0.48\textwidth}
    \includegraphics[width=\linewidth]{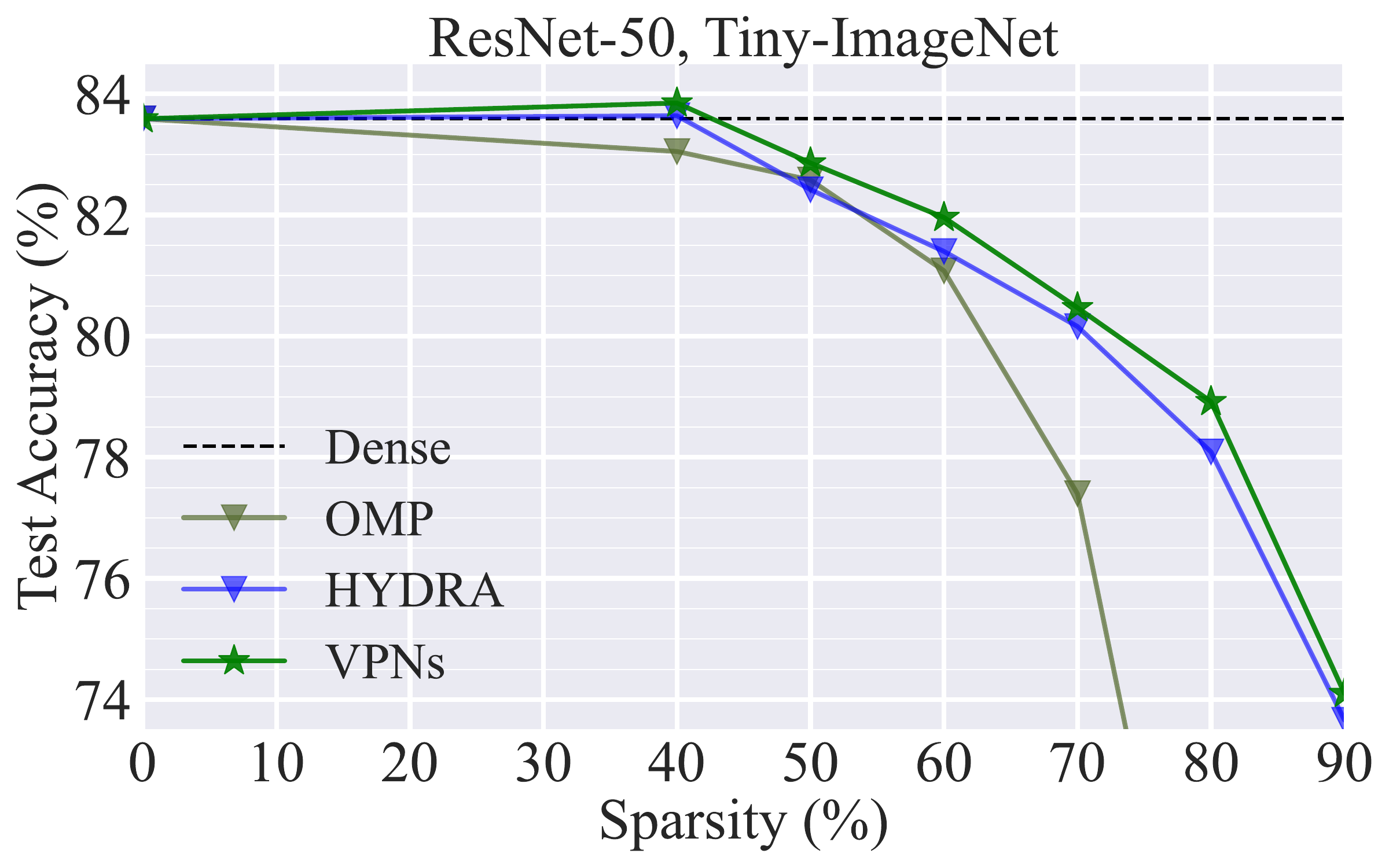}
  \end{subfigure}
  \hspace{0.1em}
  \begin{subfigure}{0.48\textwidth}
    \includegraphics[width=\linewidth]{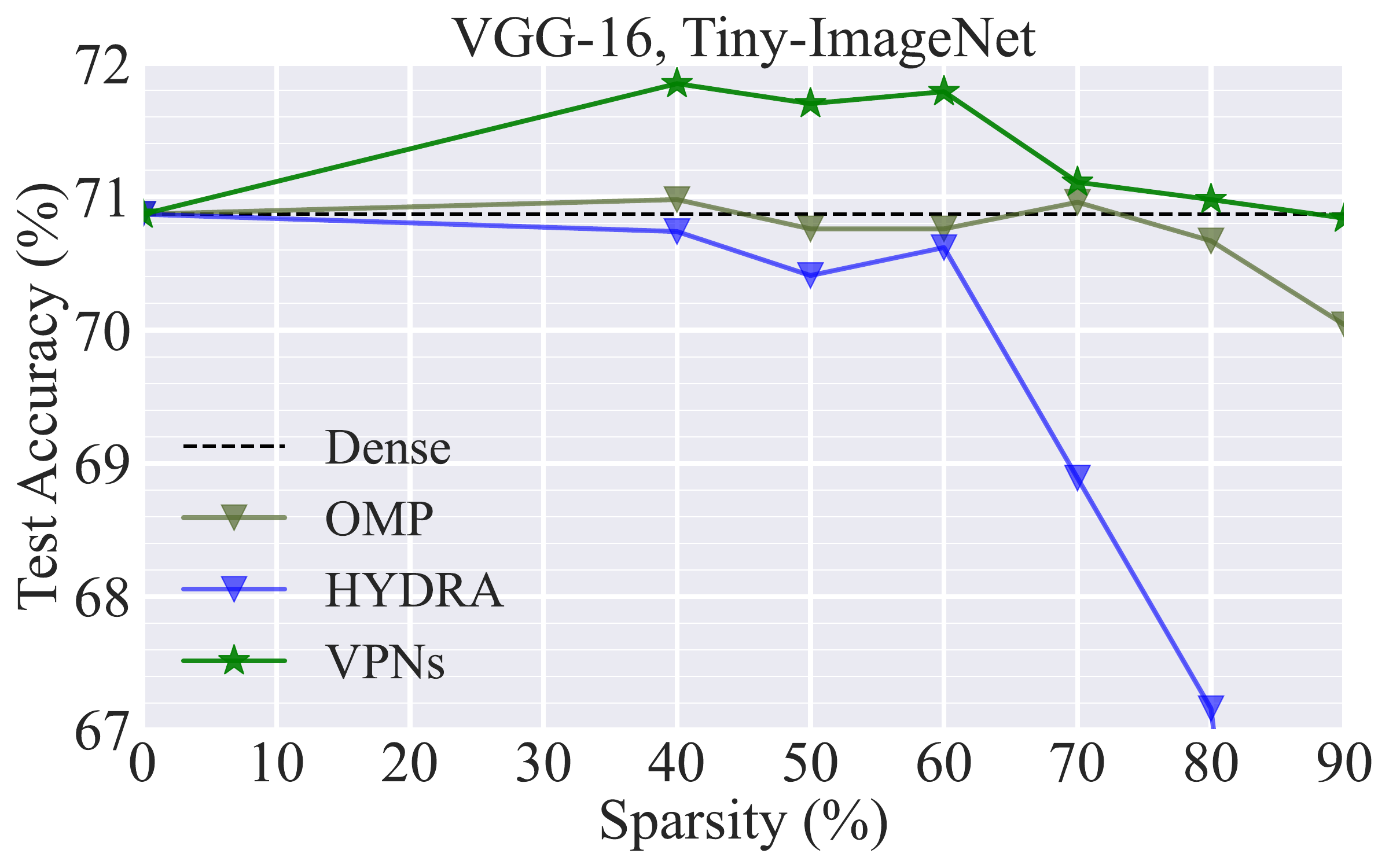}
  \end{subfigure}
  \vspace{-3mm}
  \caption{\small{\textbf{Downstream Fine-tuning Results.} The performance overview of \texttt{VPNs}, HYDRA, and OMP. All the results are obtained with ImageNet-$1$K pre-trained ResNet-$50$ and VGG-$16$, fine-tuned on Tiny-ImageNet. \texttt{VPNs} consistently has superior performance.}}
  \vspace{-3mm}
  \label{Fig. Architecture Transfer}
\end{figure}

We conduct additional experiments with ResNet-$50$ and VGG-$16$ to investigate the performance of \texttt{VPNs} over different architectures. These models are pre-trained on ImageNet-$1$K and fine-tuned on Tiny-ImageNet. All pruning methods are applied in the fine-tuning stage.  As shown in Figure~\ref{Fig. Architecture Transfer}, \texttt{VPNs} reaches outstanding performance across diverse architectures consistently, compared to OMP ($0.85\%\sim 12.23\%$ higher accuracy on ResNet-$50$) and HYDRA ($1.14\%\sim 4.08\%$ higher accuracy on VGG-$16$). It's noteworthy to highlight that OMP and HYDRA represent the most prominent baselines according to the results from Figure~\ref{Fig. Superior Performance}.


\begin{figure}[t]
  \centering
  \begin{subfigure}{0.48\textwidth}
    \includegraphics[width=\linewidth]{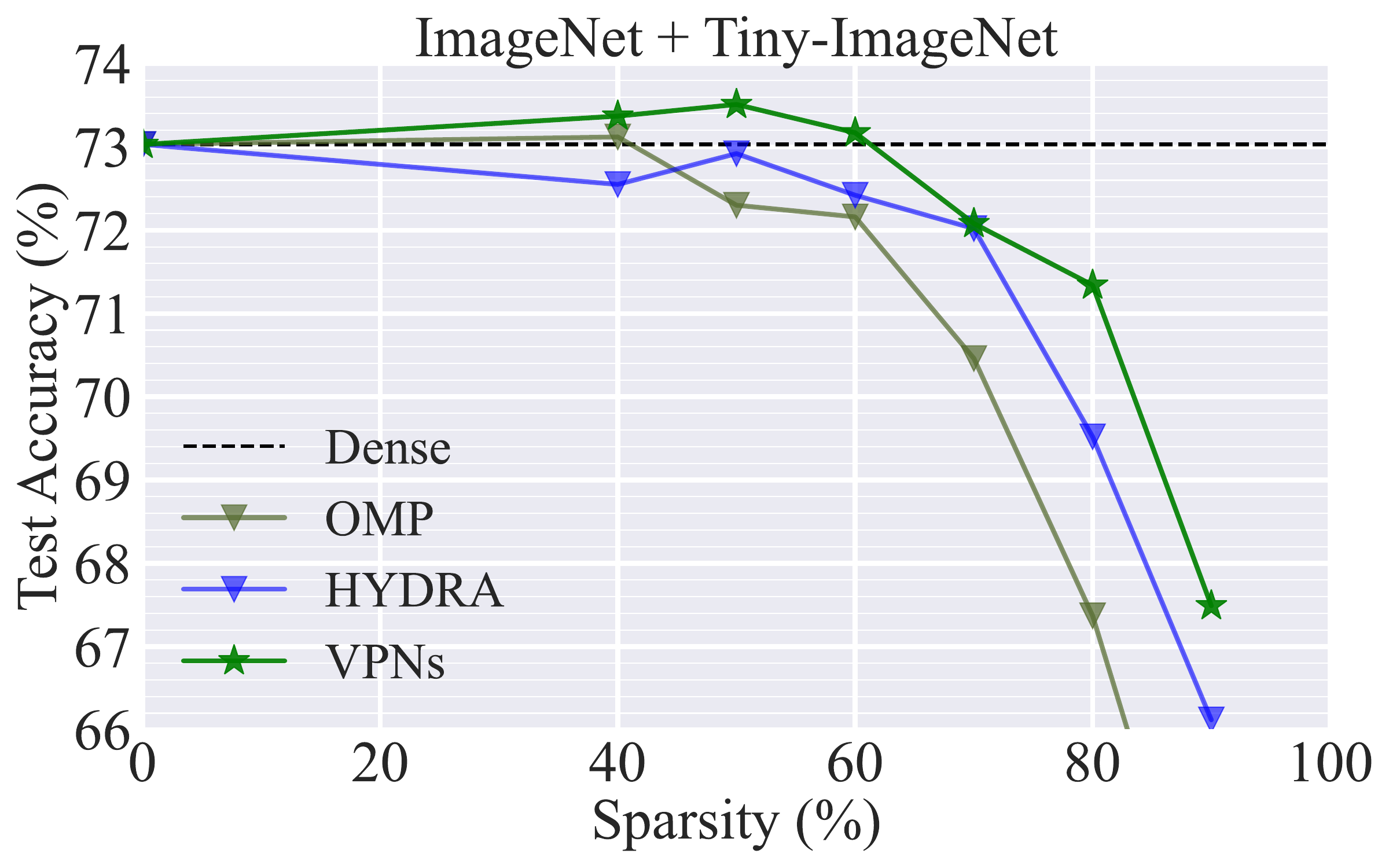}
  \end{subfigure}
  \hspace{0.1em}
  \begin{subfigure}{0.48\textwidth}
    \includegraphics[width=\linewidth]{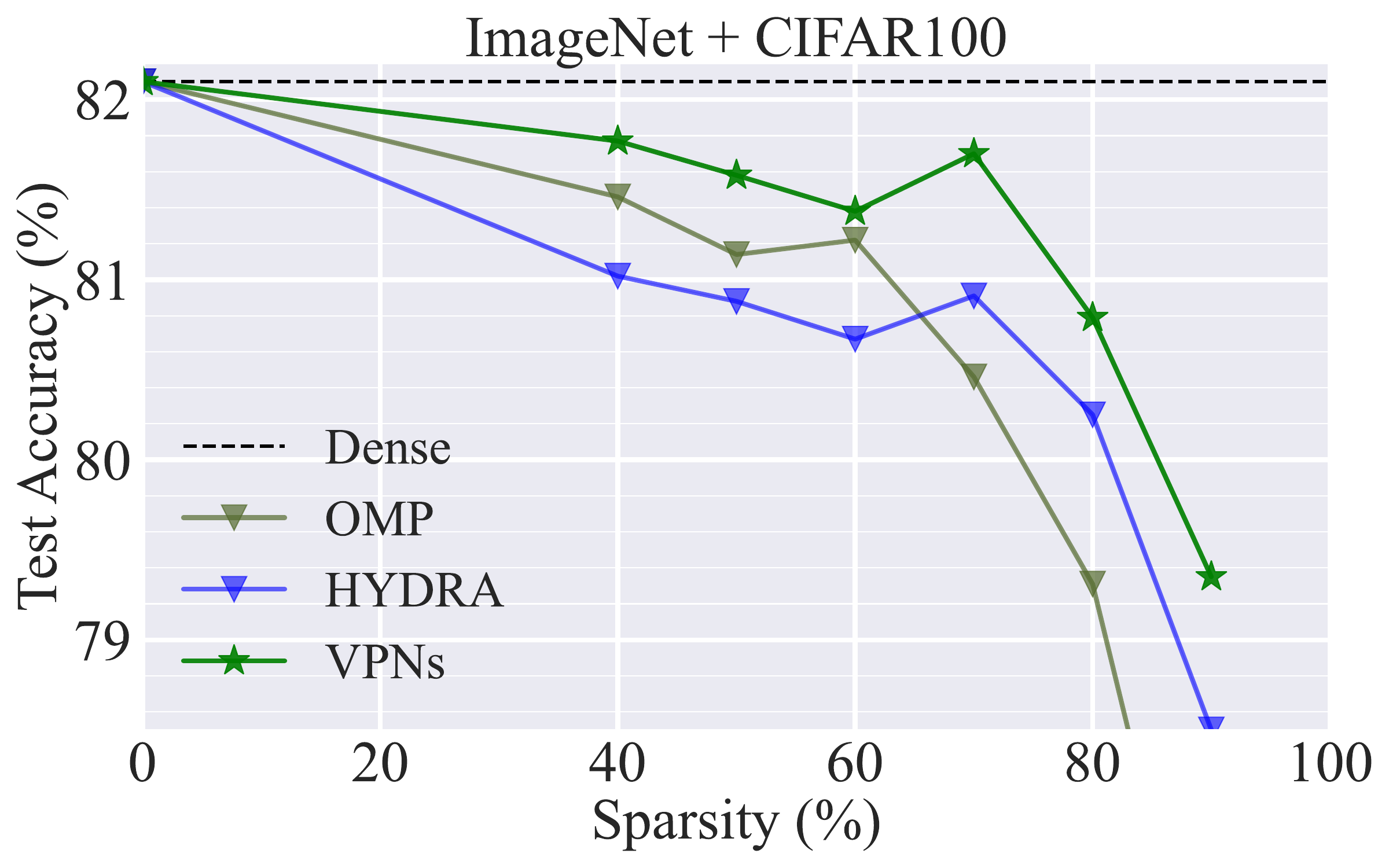}
  \end{subfigure}
  \vspace{-3mm}
  \caption{\small{\textbf{ImagetNet Mask Finding and Downstream Subnetwork Tuning Results.} The performance overview of \texttt{VPNs}, HYDRA, and OMP. The models are pruned on ImageNet-$1$K and fine-tuned on Tiny-ImageNet and CIFAR$100$. \texttt{VPNs}' subnetworks consistently enjoy the best performance which indicates \texttt{VPNs} has transferability over datasets.}}
  \label{Fig. Dataset Transfer}
  \vspace{-5mm}
\end{figure}

\vspace{-2mm}
\paragraph{Transferability of \texttt{VPNs}.}
Meanwhile, we investigate the transferability of subnetworks identified by \texttt{VPNs} across diverse downstream tasks. We apply \texttt{VPNs}, HYDRA, and OMP pruning on ResNet-$18$ and ImageNet-$1$K to identify subnetworks, subsequently fine-tune them on CIFAR$100$ and Tiny-ImageNet separately. The results are depicted in Figure \ref{Fig. Dataset Transfer}, it can be observed that: \ding{182} \texttt{VPNs} consistently excels over SoTA algorithms across multiple datasets. At an $80\%$ sparsity level on Tiny-ImageNet, \texttt{VPNs} has \{$3.97\%$, $1.57\%$\} higher test accuracy than \{OMP, HYDRA\}. Moreover, \texttt{VPNs} outperforms \{OMP, HYDRA\} by \{$2.75\%$, $0.80\%$\} at $90\%$ sparsity on CIFAR$100$. \ding{183} \texttt{VPNs} subnetworks can surpass the dense network on specific datasets. At $60\%$ sparsity on Tiny-ImageNet, subnetworks identified by \texttt{VPNs} have better performance than their dense counterparts. Consequently, \texttt{VPNs} has transferability over datasets.

\begin{figure}[t]
\vspace{-2mm}
  \centering
  \begin{subfigure}{0.32\textwidth}
    \includegraphics[width=\linewidth]{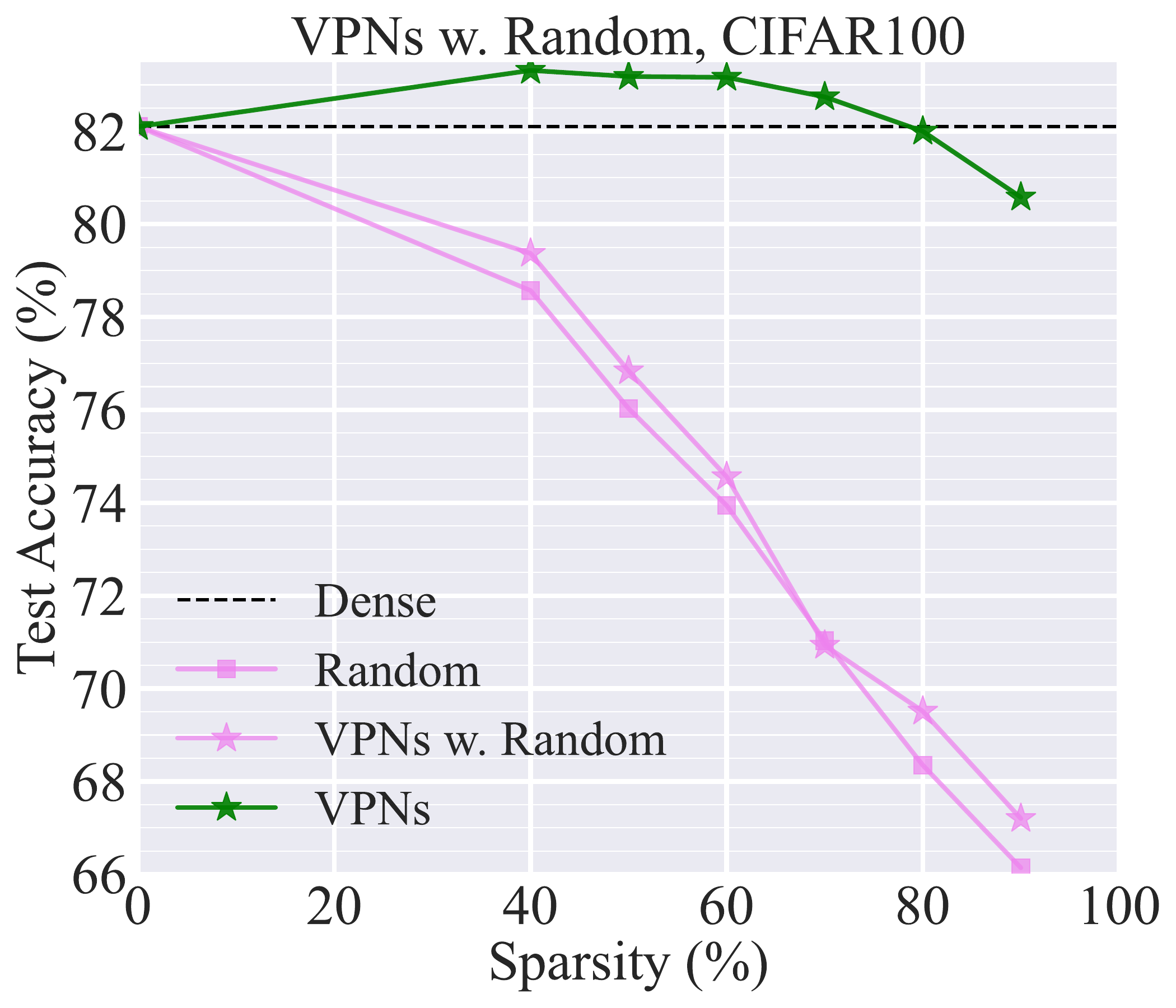}
  \end{subfigure}
  \begin{subfigure}{0.32\textwidth}
    \includegraphics[width=\linewidth]{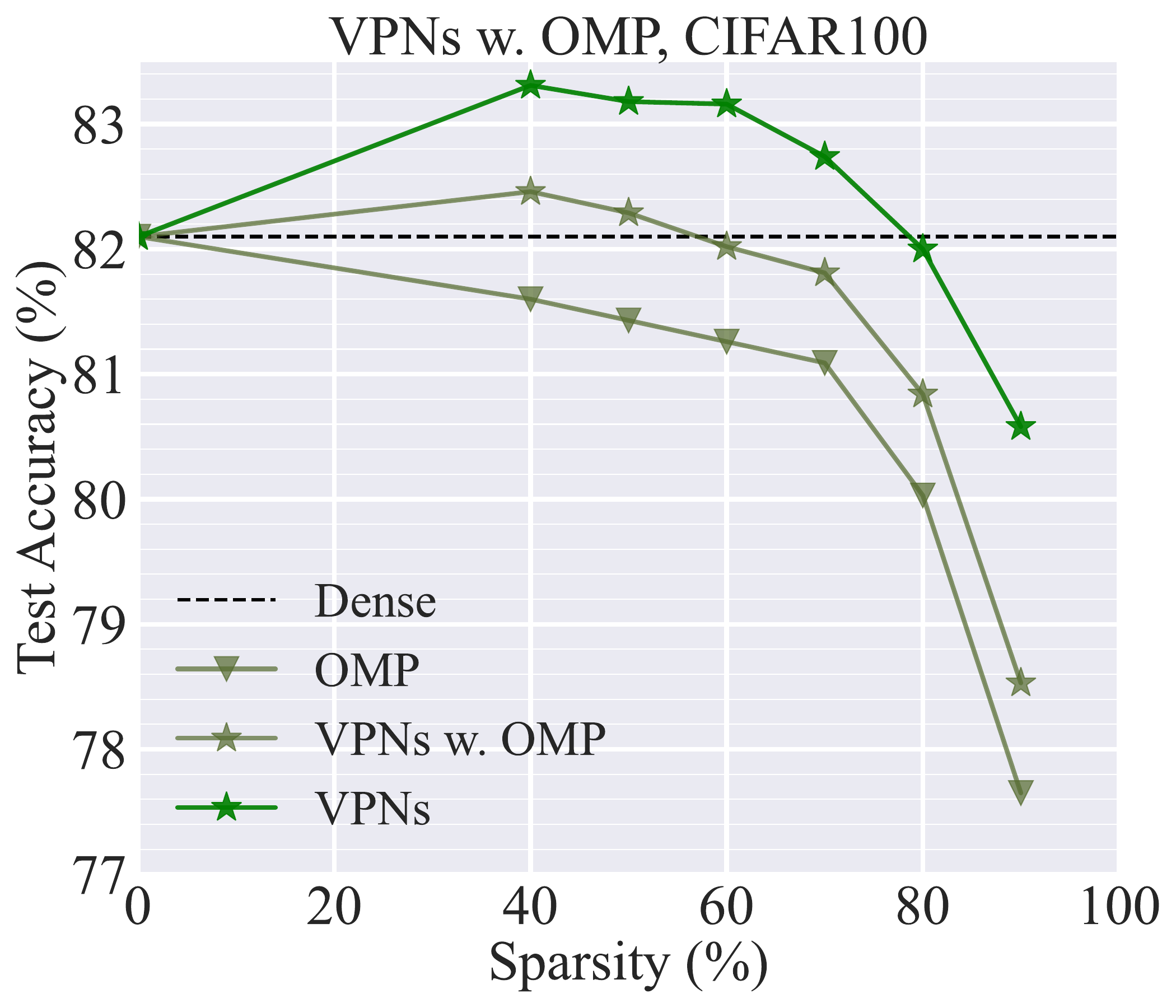}
  \end{subfigure}
  \begin{subfigure}{0.32\textwidth}
    \includegraphics[width=\linewidth]{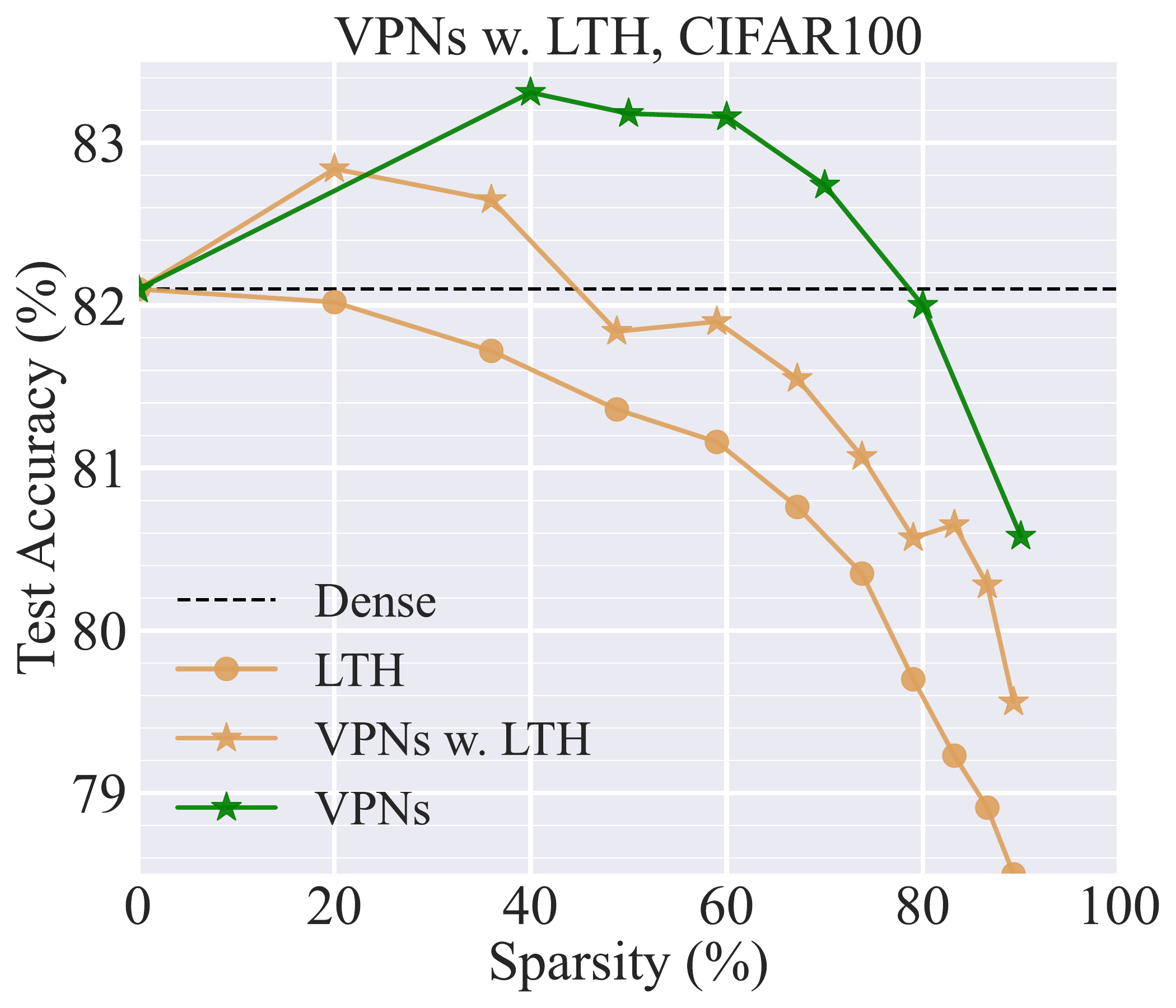}
  \end{subfigure}
  \vspace{-3mm}
  \caption{\small{\textbf{\texttt{VPNs} Paradigm Applied to Current Methods.} The performance overview of \texttt{VPNs} pruning paradigm applied to Random, OMP, and LTH pruning named \texttt{VPNs} w. Random, \texttt{VPNs} w. OMP, and \texttt{VPNs} w. LTH. The results are based on ResNet-$18$ pre-trained on ImageNet-$1$K and fine-tuned on CIFAR$100$. \texttt{VPNs} paradigm advances Random, OMP, and LTH consistently.}}
  \vspace{-2mm}
  \label{Fig. Algorithm Transfer}
\end{figure}

\vspace{-2mm}
\paragraph{Superiority of \texttt{VPNs} Pruning Paradigm.}
Furthermore, we endeavor to explore the potential of the \texttt{VPNs} pruning paradigm to enhance the effect of existing pruning algorithms. We integrate the \texttt{VPNs} pruning paradigm with Random, OMP, and LTH pruning, forming \texttt{VPNs} w. Random, \texttt{VPNs} w. OMP, and \texttt{VPNs} w. LTH respectively.  For the purpose of consistency, the VP utilized in the experiment is kept identical to the one used in \texttt{VPNs}. The results are based on ResNet-18 pre-trained on ImageNet-$1$K and fine-tuned on CIFAR$100$. As depicted in Figure \ref{Fig. Algorithm Transfer}.  We observe that \texttt{VPNs} combined with existing prunings consistently surpasses their original counterpart. For example, At $80\%$ sparsity, \{\texttt{VPNs} w. Random, \texttt{VPNs} w. OMP, \texttt{VPNs} w. LTH\} surpass their corresponding original pruning by \{$1.16\%$, $0.81\%$, $0.79\%$\}. 

\vspace{-1mm}
\subsection{Additional Investigation and Ablation Study.}
\vspace{-1mm}
\begin{figure}
  \centering
  \begin{minipage}{0.48\textwidth}
    \includegraphics[width=\linewidth]{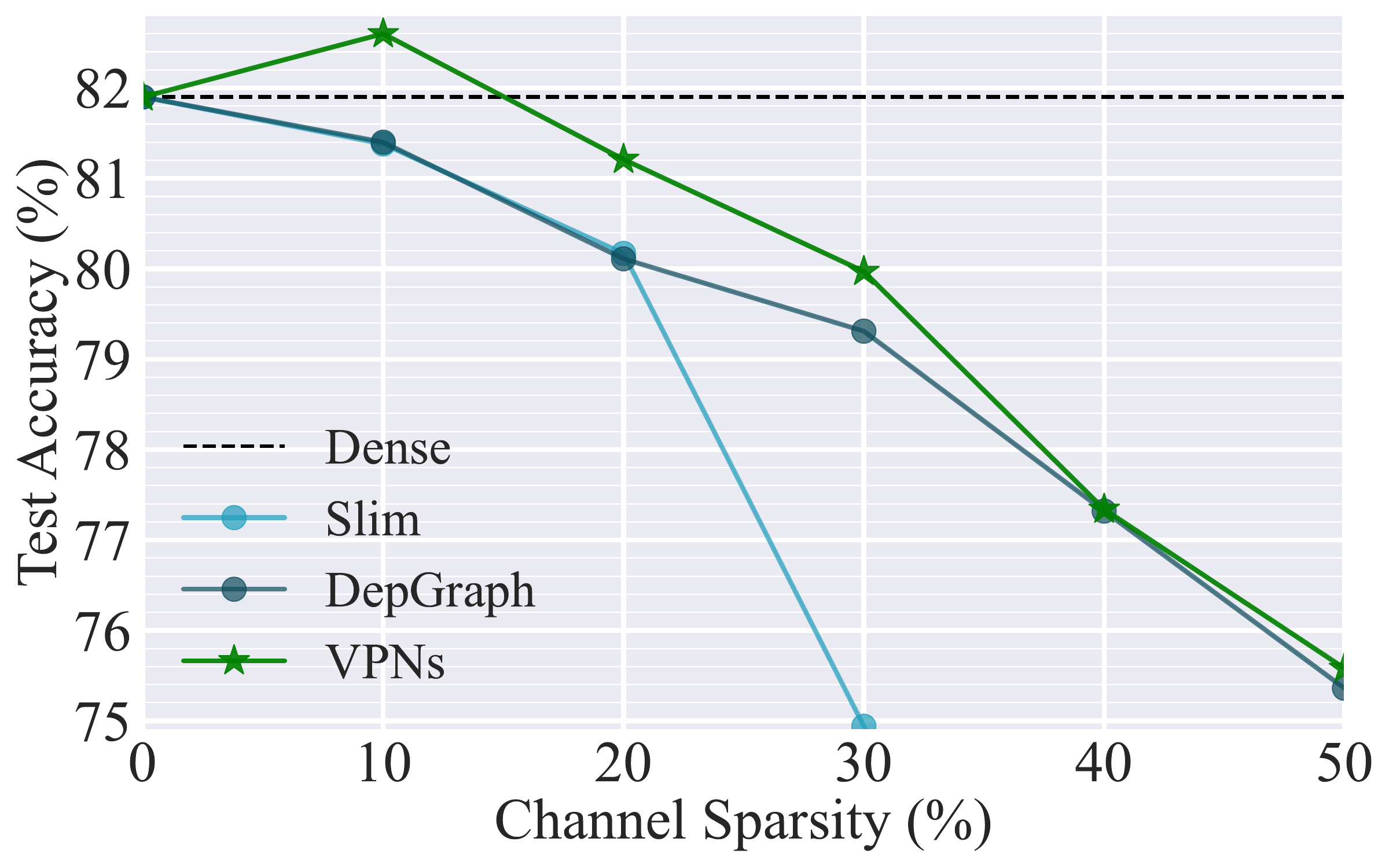}
    \vspace{-7mm}
    \caption{\small{\textbf{Structured Pruning Results.} Test accuracy of \texttt{VPNs} channel-wise pruning compared to SoTA channel-wise pruning methods on ImageNet pre-trained ResNet-$18$ and fine-tuned on CIFAR$100$. \texttt{VPNs} structured pruning has the best performance.}}
    \label{Fig. Structural Pruning}
  \end{minipage}
  \hspace{0.1em}
  \begin{minipage}{0.48\textwidth}
    \includegraphics[width=\linewidth]{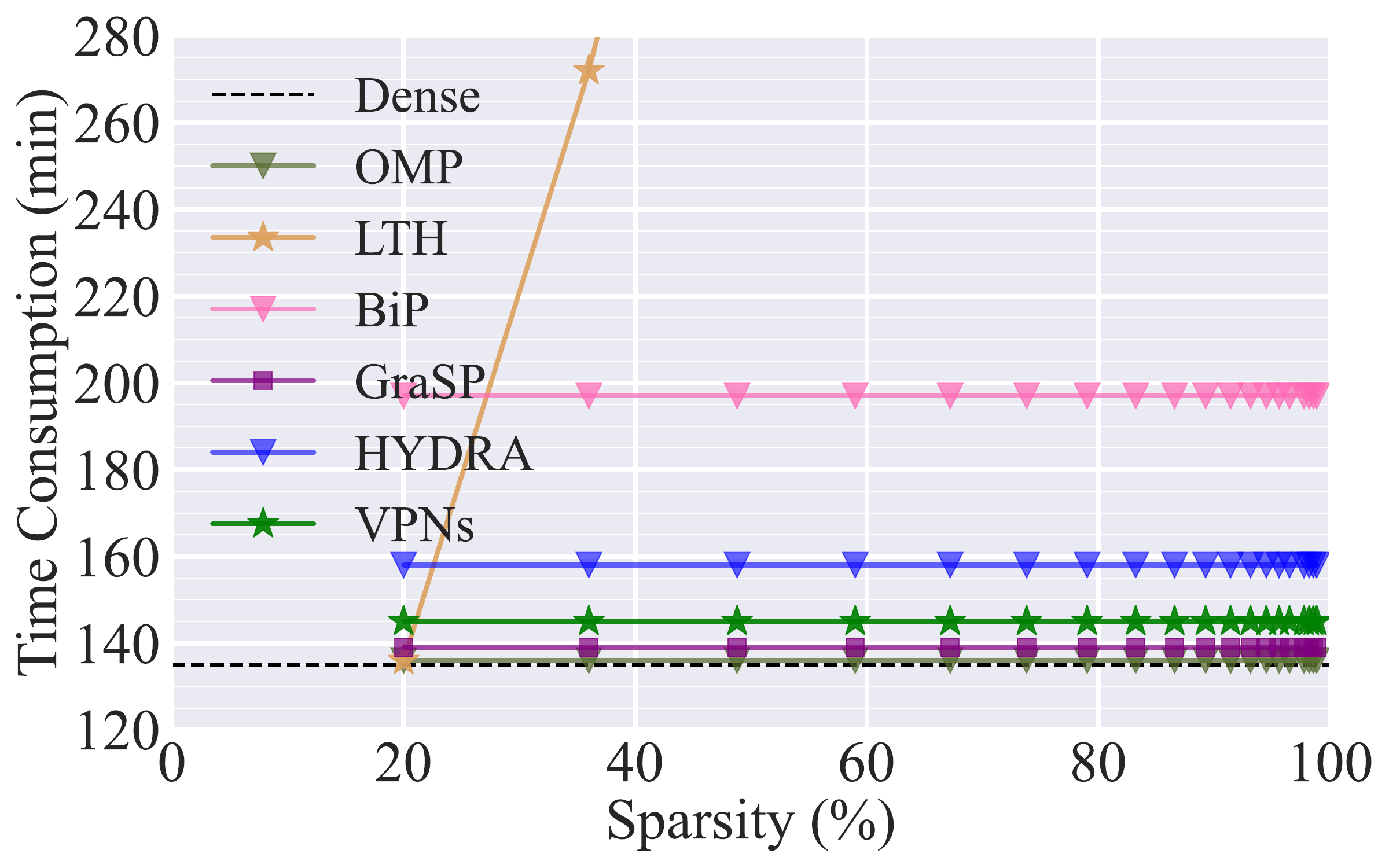}
    \vspace{-7mm}
    \caption{\small{\textbf{Time Consumption.} Time consumption of six pruning methods on ImageNet pre-trained ResNet-$18$ and fine-tuned on CIFAR$100$. \texttt{VPNs} is more time-efficient than BiP and HYDRA and nearly the same as GraSP.}}
    \label{Fig. Time}
  \end{minipage}
  \vspace{-5mm}
\end{figure}

\paragraph{\texttt{VPNs} for Structured Pruning.} To assess the potential of \texttt{VPNs} in structured pruning, we perform an empirical comparison between \texttt{VPNs} and renowned structured pruning techniques such as Slim \citep{liu2017learning} and DepGraph \citep{fang2023depgraph}. The evaluations are conducted using a pre-trained ResNet-18 model on ImageNet-1K, fine-tuned on CIFAR-100. See Appendix~\ref{Appendix: Structued pruning} for more details. From the results presented in Figure \ref{Fig. Structural Pruning}, we observe that: \ding{182} \texttt{VPNs} enjoys superior performance consistently across various channel sparsity levels in comparison to Slim and DepGraph, achieving higher accuracy by $1.04\%\sim9.54\%$ and $0.02\%\sim1.20\%$ respectively.  \ding{183} \texttt{VPNs} simultaneously reduces both training and inference FLOPs and memory costs. For example, at $10\%$ and $20\%$ channel sparsity levels, \texttt{VPNs} achieves speedup ratios of $1.1\times$ and $1.3\times$ while reducing memory costs by $15.26\%$ and $31.71\%$ respectively, without compromising the performance relative to the dense network.  The speedup ratio is quantified as $\frac{\text{FLOPs(dense)}}{\text{FLOPs(subnetwork)}}$.

\vspace{-2mm}
\paragraph{Computational Complexity.}
An effective pruning algorithm should exhibit computational efficiency. Accordingly, we evaluate the computational complexity of \texttt{VPNs} in comparison to the SoTA pruning methods. Our criterion contains training time consumption, training epochs, and gradient calculating steps with evaluations conducted on ImageNet-$1$K pre-trained ResNet-$18$ and fine-tuned on CIFAR$100$. Results are displayed in Figure \ref{Fig. Time} and Table \ref{Tab. Epochs and Steps}, several positive findings can be drawn: \ding{182} \texttt{VPNs} consistently outperforms both BiP and HYDRA in terms of time efficiency, achieving a time reduction of $26\%$ and $8.97\%$ respectively across varying sparsity levels while exhibits a time consumption comparable to GraSP. It is also noteworthy to mention that LTH's time consumption exhibits an exponential increase in relation to sparsity growth. \ding{183} \texttt{VPNs} requires the fewest epochs and steps to attain optimal performance. Specifically, for achieving a $90\%$ sparsity level, \texttt{VPNs} requires $95\%$, $50\%$, and $50\%$ fewer epochs in comparison to LTH, GraSP, and HYDRA, respectively. Moreover, it demands $90\%$ and $33\%$ fewer steps than LTH and BiP separately.


\begin{figure}[t]
\vspace{-2mm}
  \centering
  \begin{subfigure}{0.31\textwidth}
    \includegraphics[width=\linewidth]{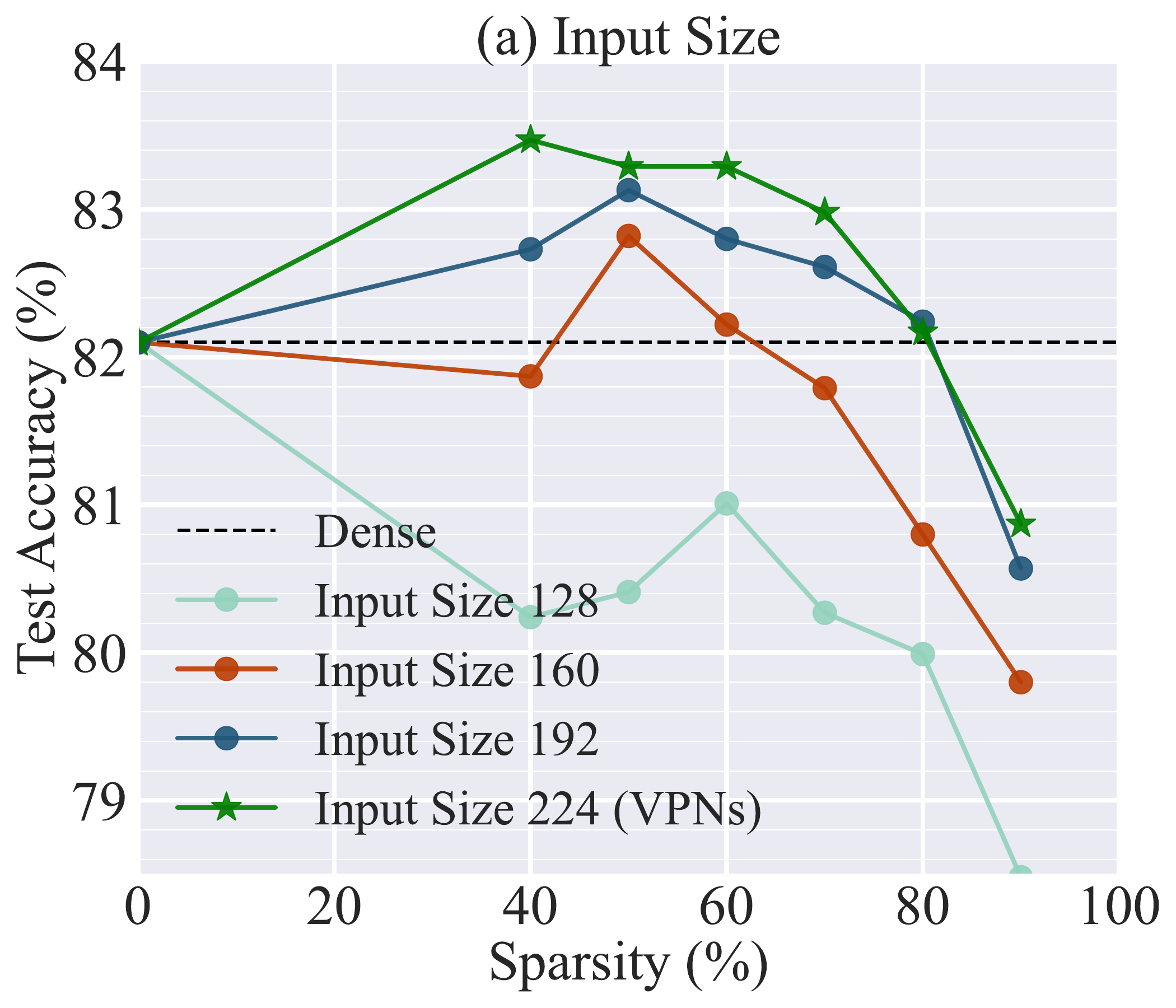}
    \phantomsubcaption
    \label{Fig. Input Size}
  \end{subfigure}
  \hspace{0.1cm}
  \begin{subfigure}{0.31\textwidth}
    \includegraphics[width=\linewidth]{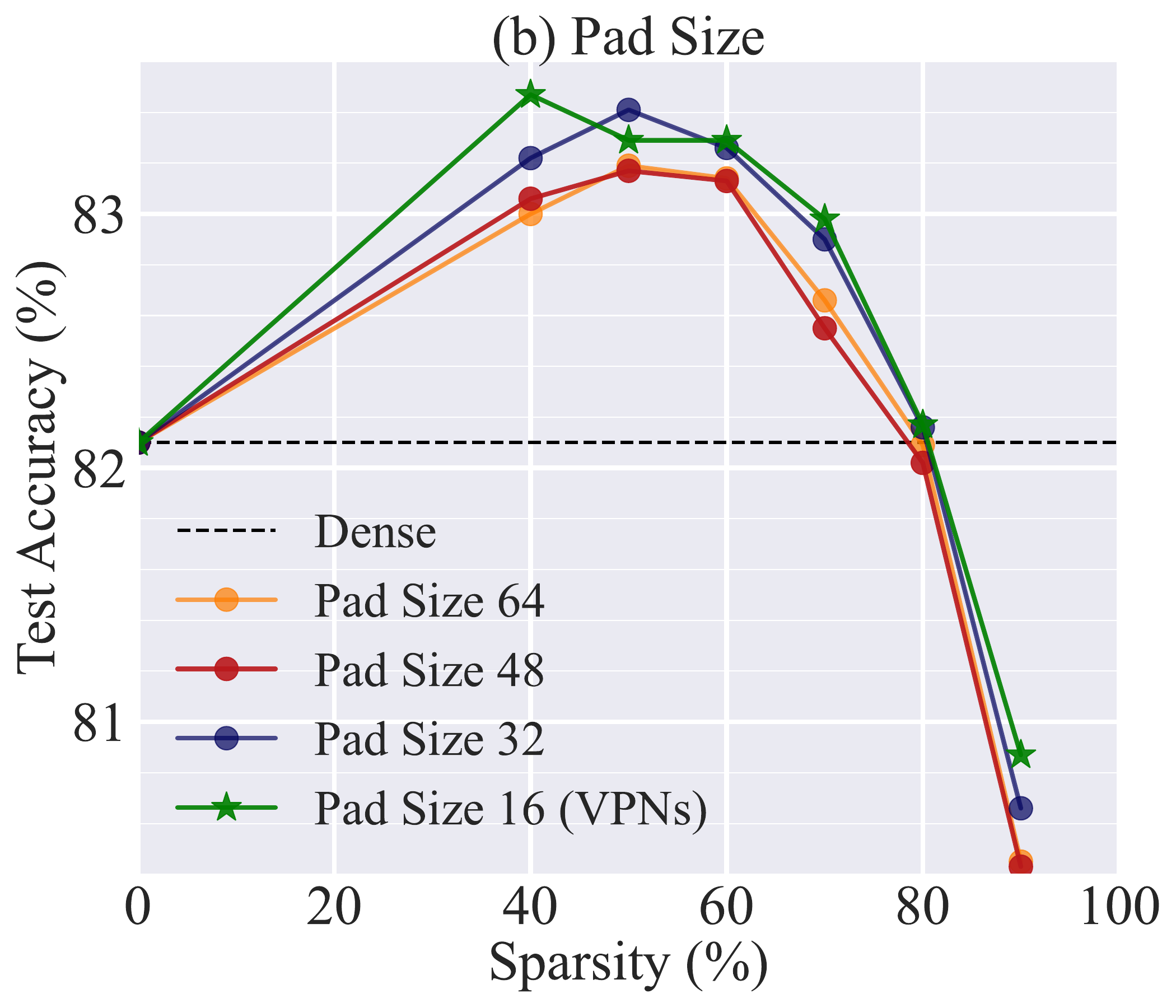}
    \phantomsubcaption
    \label{Fig. Pad Size}
  \end{subfigure}
  \hspace{0.1cm}
  \begin{subfigure}{0.31\textwidth}
    \includegraphics[width=\linewidth]{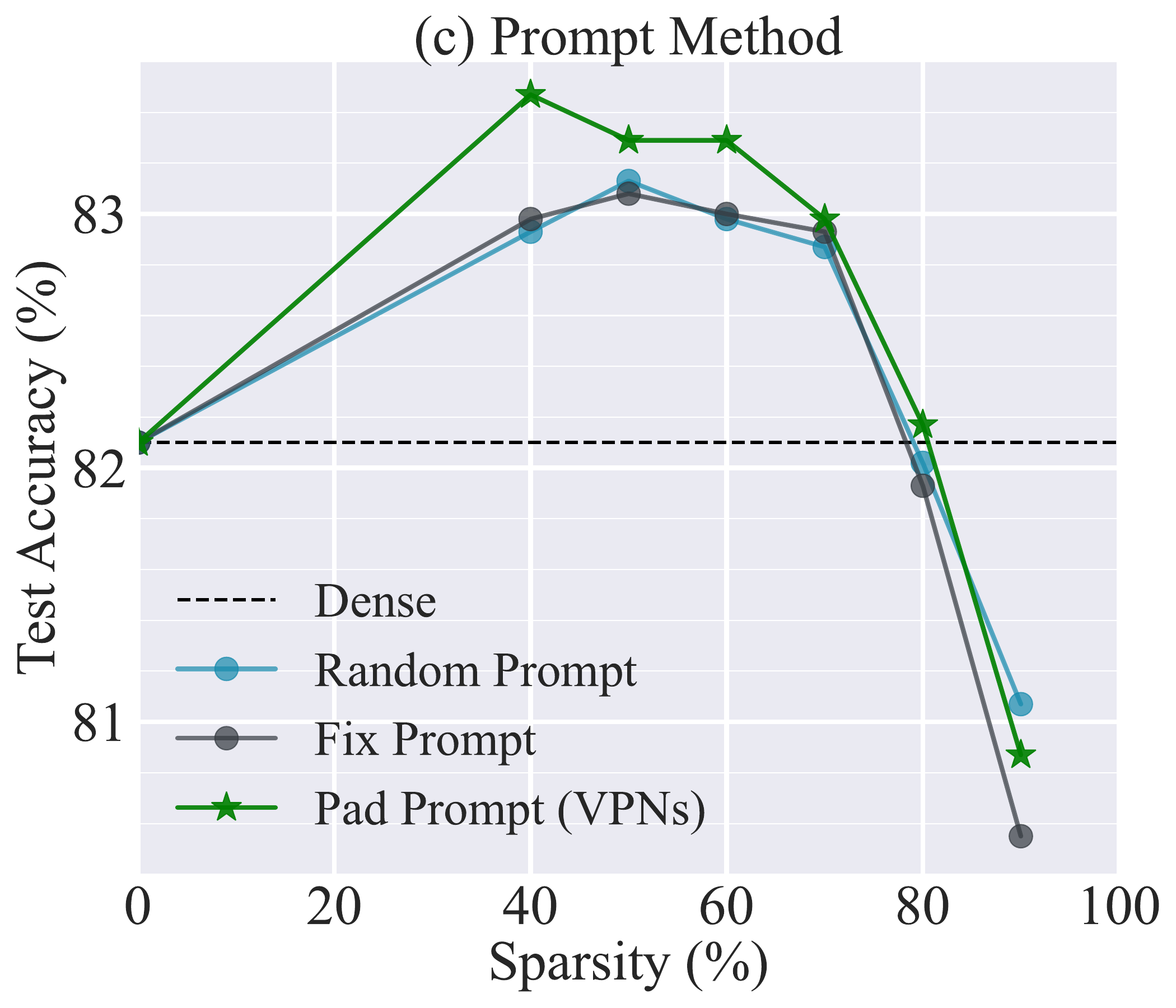}
    \phantomsubcaption
    \label{Fig. Prompt Method}
  \end{subfigure}
  \vspace{-7mm}
  \caption{\small{\textbf{Ablation of VP Designs.} Ablation studies of different VP designs on ImageNet pre-trained ResNet-$18$ and fine-tuned on CIFAR$100$. (a) Vary input size with pad prompt and pad size of $16$. (b) Vary pad size with pad prompt and input size of $224$. (c) Vary VP method with $13$K prompt parameters.}}
  \vspace{-5mm}
\end{figure}

\vspace{-2mm}
\paragraph{Ablation -- VP Designs.}
In this section, we systematically examine the impact of different VP designs on the performance of  \texttt{VPNs}. Our experiments are based on ImageNet pre-trained ResNet-$18$ and fine-tuned on CIFAR$100$, where we explore various input sizes, prompt sizes, and VP methods. 

\textit{Input Size.} We employ pad prompts with a fixed pad size of $16$ while varying the input size from $128$ to $224$ to assess the effect of input size on the performance of \texttt{VPNs}. As illustrated in Figure \ref{Fig. Input Size}, As the input size increases, we observe a corresponding rise in test accuracy. This underscores the imperative of harnessing the entirety of information available in the original images. 

\textit{Pad Size.} Similarly, to investigate the impact of the pad size, we fix the input size of $224$ and vary the pad size of the VP from  $16$ to $64$. The results are shown in~\ref{Fig. Pad Size}. Pad sizes $16$ and $32$ exhibit the best performance and the test accuracy declines as the pad sizes increase further, which indicates that a small number of prompt parameters benefits more to \texttt{VPNs} pruning performance.

\textit{Visual Prompt Strategies.} We conduct an investigation into three distinct types of VP methods: the pad prompt, the random prompt, and the fix prompt. Figure~\ref{Fig: pad Prompt} provides a visual representation of the pad prompt. In contrast, the random prompt is tunable within a randomly chosen square section of the perturbation $\bdelta$ as defined in Equation~\ref{Formula: Our Input Prompting}. The fix prompt, on the other hand, restricts tunability to the top-left square segment of  $\bdelta$. See Appendix \ref{Appendix: VP Design Details} for more details. In our experiments, all VP methods are kept consistent with 13K tunable parameters. The results are shown in Figure~\ref{Fig. Prompt Method}. We observe that the pad prompt outperforms both the fix and random prompts for \texttt{VPNs}.




\vspace{-2mm}
\paragraph{Ablation -- VP for Mask Finding/Subnetwork Tuning.}
To assess the influence of VP during the processes of mask finding and subnetwork tuning, we conduct an ablation analysis.  In this study, the VP is deactivated at specific stages of \texttt{VPNs}, resulting in two distinct algorithmic variants: `` VP for Mask Finding'' and ``VP for Subnetwork Tuning''. The results, using ResNet-18 pre-trained on ImageNet-$1$K on fine-tuned on CIFAR$100$, are depicted in Figure~\ref{Fig: VPNs phase}.  From the results, we observe that \ding{182} VP for Subnetwork Tuning contributes more to the performance gains in \texttt{VPNs} than VP for Mask Finding; \ding{183} Inserting VP in both stages of \texttt{VPNs} achieves superior test accuracy, which suggests that our proposal effectively upgrades network sparsification.


\vspace{-2mm}
\section{Conclusion}
\vspace{-1mm}

In this work, we highlight the limitations of post-pruning prompts in enhancing vision subnetworks. To harness the potential of visual prompts for vision neural network sparsification, we introduce an innovative data-model co-design algorithm, termed \textbf{\texttt{VPNs}}.  Comprehensive experiments across diverse datasets, architectures, and pruning methods consistently validate the superior performance and efficiency offered by \textbf{\texttt{VPNs}}. We further demonstrate the transferability of subnetworks identified by \textbf{\texttt{VPNs}} across multiple datasets, emphasizing its practical utility in a broader array of applications. 

\section{Reproducibility Statement}
The authors have made an extensive effort to ensure the reproducibility of the results presented in the paper. \textit{First}, the details of the experimental settings are provided in Section~\ref{sec. implementation details} and Appendix~\ref{Appendix: Implementation Details}. This paper investigates nine datasets and the details about each dataset are described in Table~\ref{Tab: Appendix Dataset Attributes}. The evaluation metrics are also clearly introduced in Section~\ref{sec. implementation details}. \textit{Second}, the baseline methods' implementation particulars are elucidated in Appendix~\ref{Appendix: Implementation Details}. Simultaneously, the implementation details of our method, \textbf{\texttt{VPNs}}, are included in Section~\ref{sec. implementation details} and Appendix~\ref{Appendix: Implementation Details}. \textit{Third}, the codes are included in the supplementary material for further reference.


\bibliography{iclr2024_conference}
\bibliographystyle{iclr2024_conference}
\newpage

\renewcommand{\thepage}{A\arabic{page}}  
\renewcommand{\thesection}{A\arabic{section}}   
\renewcommand{\thetable}{A\arabic{table}}   
\renewcommand{\thefigure}{A\arabic{figure}}
\appendix
\vspace{-3mm}
\section{\textbf{\texttt{VPNs}} Algorithm Details}\label{Appendix: VPNs Algorithm}
\vspace{-1mm}
Here we provide the pseudo-code of \textbf{\texttt{VPNs}}. It first creates prompted images, and then locates the sparse subnetwork by jointly optimizing the mask and VP. Finally, the weights of found sparse subnetwork are further fine-tuned together with the VP.
\begin{algorithm}
\caption{\textbf{\texttt{VPNs}}}
\begin{algorithmic}[1]
\Require Dataset $\mathcal{D} = \{(\bx_1, y_1),...,(\bx_n, y_n)\}$, pre-trained model $f_{\btheta_{\text{pre}}}$, and sparse ratio $s$.
\Ensure Sparse neural network $f_{\btheta_{\text{fine-tune}} \odot \bmm_s}$.
\State Input VP operation: $\bx'(\bdelta) = h(\bx, \bdelta) = r^i(\bx) + \bdelta^p, \bx \in \mathcal{D}$.
\State Sparse Initialization: Initialize importance score $\bc$ and update the mask $\bmm=\mathbb{I}(|\bc|>|\bc|_{1-s})$, where $|\bc|_{1-s}$ is the $1-s$ percentile of $|\bc|$.
\For{$i$=1 to epochs}
\State Caculate pruning loss: $\mathcal{L}_i = \mathbb{E}_{(\bx, y)\in \mathcal{D}}\mathcal{L}(f_{\btheta_{\text{pre}} \odot \bmm_{i-1}}(\bx'(\bdelta_{i-1})), y)$.
\State Update VP $\bdelta_i$ and importance scores $\bc_i$ via SGD calling with $\btheta$ frozen.
\State Update the mask: $\bmm_i=\mathbb{I}(|\bc_i|>|\bc_i|_{1-s})$.
\EndFor
\State Re-initialization: Initialize VP with $\bdelta_s$, the mask $\bmm$ with $\bmm_s$, and $\btheta$ with $\btheta_{\text{pre}}$.
\For{$j$=1 to epochs}
\State Caculate fine-tuning loss: $\mathcal{L}_j = \mathbb{E}_{(\bx, y)\in \mathcal{D}}\mathcal{L}(f_{\btheta_{j-1} \odot \bmm_s}(\bx'(\bdelta_{j-1})), y)$.
\State Update VP $\bdelta_j$ and model weights $\btheta_j$ via SGD calling with $\bmm$ frozen.
\EndFor
\end{algorithmic}
\label{Algo: VPNs}
\end{algorithm}
\vspace{-3mm}
\section{Visual Prompt Design Details}\label{Appendix: VP Design Details}
\vspace{-1mm}
We explore three different kinds of VP designs, namely \textbf{Pad Prompt}, \textbf{Random Prompt}, and \textbf{Fix Prompt} \citep{bahng2022exploring}. Each of these VP methods can be formulated into two steps: \ding{182} \textit{Input resize and pad operation.} We resize the original image $\bx$ to a designated input size $i\times i$ and subsequently pad it to $224\times224$ with $0$ values to derive the resized image. This procedure is represented as $r^i(\bx)$, where $r(\cdot)$ refers to the resize and pad operation and $i$ indicates the input size. \ding{183} \textit{Perturbation mask operation.} We initiate the perturbation parameters of $\bdelta$ as a $224\times224$ matrix with a portion being masked. The input prompting operation is then formulated as Equation~\ref{Formula: Our Input Prompting}. All the VP variants have the same input resize and pad operation and the differences for them lie in the distinct masked regions during the perturbation mask operation.

\textit{Pad Prompt.} This kind of prompt masks a central square matrix of the perturbation $\bdelta$, while keeping the left four peripheral segments tunable. The width of each side is denoted as the pad size, marked as $p$. Figure~\ref{Fig: pad Prompt} provides a visual representation of the pad prompt. The number of tunable prompt parameters numbers for the pad prompt is $4p(224-p)$.

\textit{Fix Prompt.}  This prompt design retains the top-left square segment of the perturbation $\bdelta$ tunable, masking the remaining areas of $\bdelta$. The width of the tunable square is denoted prompt size, marked as $p$. The number of tunable prompt parameters for the fix prompt is $p^2$.

\textit{Random Prompt.} The random prompt keeps a random square segment of the perturbation $\bdelta$ tunable, masking other areas of $\bdelta$ during each forward pass. Similarly, the width of the tunable square is denoted as $p$ and referred to as the prompt size. The random prompt has a $p^2$ parameter number. 

\begin{table}[H]
\vspace{-1mm}
\centering
\caption{\small{Datasets configurations.}}
\vspace{-5mm}
\label{Tab: Appendix Dataset Attributes}
\begin{center}
\begin{tabularx}{0.85\textwidth}{Xcccc}
\toprule
Dataset & Train Set Size & Test Set Size & Class Number & Batch Size  \\ 
\midrule
Flowers$102$ & $5726$ & $2463$ & $102$ & $128$   \\
DTD & $3760$ & $1880$ & $47$ & $64$   \\
Food$101$ & $75750$ & $25250$ & $101$ & $256$   \\
OxfordPets & $3680$ & $3669$ & $37$ & $64$   \\
StanfordCars & $8144$ & $8041$ & $196$ & $128$   \\
CIFAR$10$ & $50000$ & $10000$ & $10$ & $256$   \\
CIFAR$100$ & $50000$ & $10000$ & $100$ & $256$   \\
Tiny ImageNet & $100000$ & $10000$ & $200$ & $256$   \\
ImageNet & $1281167$ & $50000$ & $1000$ & $1024$   \\
\bottomrule
\end{tabularx}
\vspace{-4mm}
\end{center}
\end{table}
\section{Implementation Details}\label{Appendix: Implementation Details}
\vspace{-1mm}
\paragraph{Datasets.}\label{Appendix: Datasets.}
We use the standard train-test division of 9 image classification datasets to implement our method and report the test set accuracy. All images are resized to $224\times 224$ in mask finding and weight tuning processes. The configurations of the datasets are summarized in Table \ref{Tab: Appendix Dataset Attributes}.

\begin{wrapfigure}{r}{0.4\textwidth}
\centering
    \vspace{-5mm}
    \includegraphics[width=\linewidth]{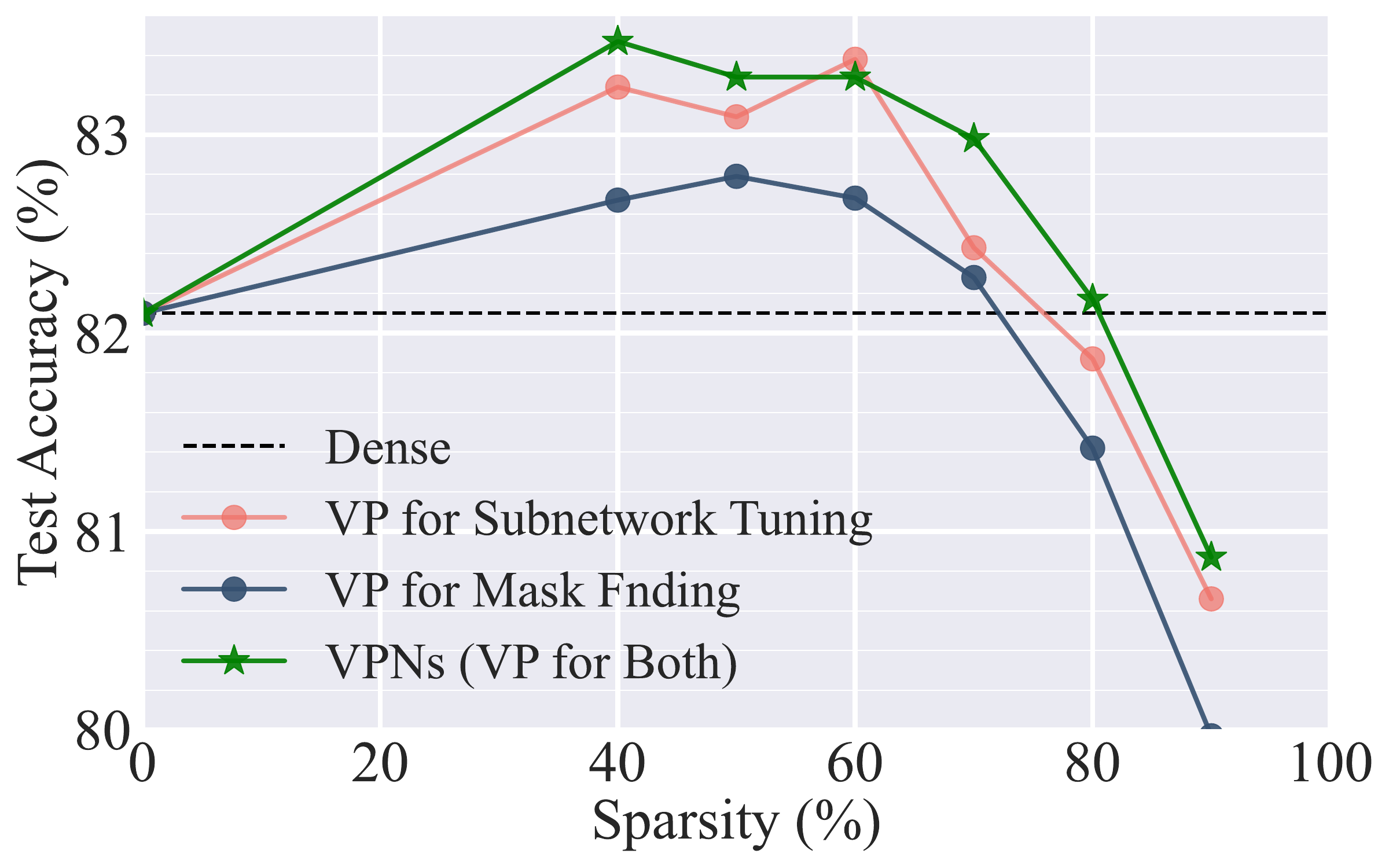}
    \vspace{-7mm}
    \caption{\small{\textbf{Ablation of VP for Mask Finding/Subnetwork Tuning.} Ablation of \texttt{VPNs} only using VP in mask finding or subnetwork tuning on ImageNet pre-trained ResNet-$18$ and fine-tuned on CIFAR$100$.}}
    \label{Fig: VPNs phase}
    \vspace{-8mm}
\end{wrapfigure}

\paragraph{Hyperparameters for Unstructured Pruning.}\label{Appendix: Unstructured Pruning.}
For Random~\citep{liu2022unreasonable}, OMP~\citep{han2015deep}, LTH~\citep{chen2021lottery}, SNIP~\citep{lee2018snip}, GraSP~\citep{wang2020gradient}, SynFlow~\citep{tanaka2020pruning}, we use SGD optimizer with a learning rate of $0.01$ and cosine decay scheduler. For HYDRA~\citep{sehwag2020hydra}, BiP~\citep{zhang2022advancing} and \texttt{VPNs}, we use Adam optimizer with a learning rate of $0.0001$ and cosine decay scheduler for mask finding and SGD optimizer with a learning rate of $0.01$ and cosine decay scheduler for weight tuning. Further details regarding hyperparameter configurations can be found in Table~\ref{Tab: Appendix Hyperparameters and Configurations}. In this work, we use $\bmm$ to indicate hyperparameters for mask finding and $\btheta$ represents hyperparameters for weight tuning.

\begin{table}[t]
\vspace{-2mm}
\centering
\caption{\small Configurations for unstructured pruning. $\bmm$ indicates hyperparameters for maskfinding and $\btheta$ represents hyperparameters for weight tuning.}
\vspace{-5mm}
\label{Tab: Appendix Hyperparameters and Configurations}
\begin{center}
\resizebox{1\textwidth}{!}{
\begin{tabular}{cccccc}
\toprule
Method & Epochs & Optimizer & Initial Learning Rate & Learning Rate Decay & Weight Decay  \\ 
\midrule
Random & $120$ & SGD & $0.01$ & Cosine Decay & $0.0001$   \\
OMP & $120$ & SGD & $0.01$ & Cosine Decay & $0.0001$  \\
LTH & $120$ & SGD & $0.01$ & Cosine Decay & $0.0001$  \\
SNIP & $120$ & SGD & $0.01$ & Cosine Decay & $0.0001$  \\
GraSP & $120$ & SGD & $0.01$ & Cosine Decay & $0.0001$  \\
SynFlow & $120$ & SGD & $0.01$ & Cosine Decay & $0.0001$  \\
BiP & $60$  & Adam for $\bdelta$, SGD for $\btheta$ & $0.0001$ for $\bdelta$, $0.01$ for $\btheta$ & Cosine Decay & $0.0001$   \\
HYDRA & $60$ for $\bdelta$, $60$ for $\btheta$  & Adam for $\bdelta$, SGD for $\btheta$ & $0.0001$ for $\bdelta$, $0.01$ for $\btheta$ & Cosine Decay & $0.0001$   \\
\textbf{\texttt{VPNs}} & $30$ for $\bdelta$, $30$ for $\btheta$  & Adam for $\bdelta$, SGD for $\btheta$ & $0.0001$ for $\bdelta$, $0.01$ for $\btheta$ & Cosine Decay & $0.0001$   \\
\bottomrule
\end{tabular}}
\end{center}
\vspace{-1mm}
\end{table}

\paragraph{Hyperparameters for Structured Pruning.}\label{Appendix: Structued pruning}
We follow the implementation in \citet{fang2023depgraph} to reproduce the results of Slim \citep{liu2017learning} and DepGraph \citep{fang2023depgraph}. For the structured pruning version of \textbf{\texttt{VPNs}}, we warm up $5$ epochs before pruning. In the mask finding stage, we train $30$ epochs using the Adam optimizer with a learning rate of $1$ and cosine decay scheduler.  The weight decay is set to $0.01$. In the weight tuning stage, we train $30$ epochs using the SGD optimizer with a learning rate of $0.01$, cosine decay scheduler, and weight decay of $0.0001$. 

\begin{table}[t]
  \vspace{-2mm}
  \centering
  \captionof{table}{\small{\textbf{Training Epochs and Steps.} Training epochs and gradients calculating steps among different pruning algorithms on ImageNet pre-trained ResNet-$18$ and fine-tuned on CIFAR$100$. \texttt{VPNs} takes the least epochs and steps to obtain the superior performances of CIFAR$100$ in Figure \ref{Fig. Superior Performance}.}}
  \vspace{-3mm}
  \label{Tab. Epochs and Steps}
  \resizebox{1\textwidth}{!}{
    \begin{tabular}{lcccccccc}
      \toprule
      Method & \multicolumn{4}{c}{Epochs} & \multicolumn{4}{c}{Steps} \\ 
      \cmidrule(lr){2-5} \cmidrule(lr){6-9}
      Sparsity & $20\%$ & $59\%$ & $89.26\%$ & $95.60\%$ & $20\%$ & $59\%$ & $89.26\%$ & $95.60\%$ \\
      \midrule
      LTH & $120$ & $480$ & $1200$ & $1680$ & $23520$ & $94080$ & $235200$ & $329280$\\
      OMP & \multicolumn{4}{c}{$120$} & \multicolumn{4}{c}{$23520$} \\
      GraSP & \multicolumn{4}{c}{$120$} & \multicolumn{4}{c}{$23523$} \\
      BiP & \multicolumn{4}{c}{$60$} & \multicolumn{4}{c}{$34560$} \\
      HYDRA & \multicolumn{4}{c}{$60$ mask finding + $60$ subnetwork tuning} & \multicolumn{4}{c}{$11760$ mask finding + $11760$ subnetwork tuning} \\ \midrule
      \textbf{\texttt{VPNs}} & \multicolumn{4}{c}{$30$ mask finding + $30$ subnetwork tuning} & \multicolumn{4}{c}{$11760$ mask finding + $11760$ subnetwork tuning} \\
      \bottomrule
    \end{tabular}
  }
  \vspace{-2mm}
\end{table}

\vspace{-1mm}
\section{Additional Results}\label{Appendix: Addtional Results}
\vspace{-1mm}

\paragraph{Computational Complexity.} 
Here we provide additional results of computational complexity analysis among \textbf{\texttt{VPNs}} and our baselines through the lens of training epochs and gradient calculating steps. The experiments are conducted on ImageNet pre-trained ResNet-$18$ and fine-tuned on CIFAR$100$. From Table \ref{Tab. Epochs and Steps}, we observe that \textbf{\texttt{VPNs}} requires the fewest epochs and steps to attain optimal performance, which means \textbf{\texttt{VPNs}} is highly computationally efficient.

\paragraph{Ablation -- VP for Mask Finding/Subnetwork Tuning.}
Figure \ref{Fig: VPNs phase} shows the ablation results among \textbf{\texttt{VPNs}} (``VP for Both''), `` VP for Mask Finding'' and ``VP for Subnetwork Tuning''. The experiments are based on ResNet-18 pre-trained on ImageNet-$1$K on fine-tuned on CIFAR$100$. We find that 
\textbf{\texttt{VPNs}} achieves superior performance, which means inserting VP in both stages is the best.

\end{document}

%% file: math_commands.tex
\usepackage{amsmath,amsfonts,bm}

\def\eqref#1{equation~\ref{#1}}

\def\1{\bm{1}}

\DeclareMathAlphabet{\mathsfit}{\encodingdefault}{\sfdefault}{m}{sl}
\SetMathAlphabet{\mathsfit}{bold}{\encodingdefault}{\sfdefault}{bx}{n}

